\documentclass[11pt,a4paper]{article}

\ifdefined\pdfgentounicode  
  \input{glyphtounicode}
\fi
\ifdefined\pdfinterwordspaceon
  \pdfinterwordspaceon
\fi

\usepackage[utf8]{inputenc}
\usepackage{cmap}
\usepackage[T1]{fontenc}
\usepackage{lmodern}
\usepackage{microtype}
\usepackage[
    a4paper,
    top=2.5cm,
    bottom=2.5cm,
    left=2.5cm,
    right=2.5cm,
    headheight=14pt
]{geometry}
\usepackage{amsmath,amssymb}
\usepackage{graphicx}
\usepackage{float}
\usepackage{booktabs}
\usepackage{array}
\usepackage{longtable}
\usepackage{enumitem}
\usepackage{xcolor}
\usepackage[font=small,justification=raggedright,singlelinecheck=false]{caption}
\usepackage{url}
\usepackage{indentfirst}
\usepackage{flafter}
\usepackage{placeins}
\usepackage{needspace}
\usepackage[numbers,sort&compress]{natbib}
\usepackage[colorlinks=true,linkcolor=blue,citecolor=blue,urlcolor=blue]{hyperref}
\usepackage[nameinlink,capitalise,noabbrev]{cleveref}

\newcolumntype{L}[1]{>{\raggedright\arraybackslash}m{#1}}
\newcolumntype{C}[1]{>{\centering\arraybackslash}m{#1}}
\newcommand{\TableBody}{\small\renewcommand{\arraystretch}{1.20}}
\newcommand{\TableNote}{\footnotesize}
\newcommand{\TableNoteGap}{\vspace{0.65em}}
\makeatletter
\renewcommand\paragraph{\@startsection{paragraph}{4}{\z@}%
  {0.90ex}%
  {-0.75em}%
  {\normalfont\normalsize\bfseries}}
\makeatother
\newcommand{\PostTableHeadingSpace}{\FloatBarrier\vspace{0.35\baselineskip}}

\newcommand{\PostReferencesSpace}{\FloatBarrier\vspace{0.75\baselineskip}}

\title{\textbf{From Prompts to Contracts: Harness Engineering\\for Auditable Enterprise LLM Agents}\\[1em]}

\author{%
  Joongho Ahn\\
  AI Leadership Research Center\\
  \texttt{jhahn@ceoai.kr}
  \and
  Moonsoo Kim\thanks{Corresponding author.}\\
  AI Leadership Research Center\\
  \texttt{mskim@ceoai.kr}
}

\date{}

\begin{document}

\maketitle

\begin{abstract}
  Enterprise large language model (LLM) applications often begin as prototypes whose behavior is carried by prompts and retrieval context. Productization adds requirements for source boundaries, entity routing, answer contracts, and reproducible traces. We present a harness-engineering approach that reconstructs this pattern into a traceable, auditable LLM-agent architecture: deterministic behavior moves into code, manifests, schemas, and validation artifacts around a replaceable composition boundary, while source-backed claims remain the authority for runtime answers. We instantiate it on a public-data slice of five Korean corporate groups (25 listed companies) and evaluate three research questions. (1) The harness preserves its source-grounding, entity-routing, trace, output-hygiene, and recommendation-language contracts across the fixed validation scenarios; a fault-injection control confirms the validators flag deliberately broken contracts. (2) The checks the harness enforces held under model substitution: across three hosted models, they passed on all 270 composition-boundary runs; failures were confined to the model-composed side and were caught and recorded. (3) The code-owned guarantees are load-bearing, not reproducible by prompting alone: holding the model fixed and varying only the enforcement layer, prompt instructions alone let recommendation-language and internal-trace-leakage violations reach the reader, which the harness blocks entirely. A bolt-on external guardrail prevents such violations too but over-refuses, dropping utility to 88/120 where the harness preserves full utility (120/120); in this ablation, only code-owned enforcement preserves both safety and utility. The result is a reusable engineering pattern for turning exploratory prototypes into auditable applications with versioned source, control, and validation artifacts.
\end{abstract}

\section{Introduction}

Enterprise large language model (LLM) applications are often explored first as demonstrations: a system prompt, a set of retrieved documents, and a user interface that shows the intended interaction. We call a prototype prompt-dominant when important product behavior is encoded mainly as natural-language instructions or broad retrieval context, with code, data contracts, and validation artifacts still underdeveloped. Prompt-dominant development is useful for exploration and is related to vibe coding, a conversational LLM-assisted software-development workflow popularized by Karpathy \citep{karpathy2025vibecoding,meske2025vibecoding}. Such prompts carry behavior well enough to demonstrate it, but not to guarantee it. Productization introduces requirements that prompts alone do not enforce: each visible claim should be traceable to bounded sources, routed to the correct entity, constrained in what it may assert, regenerated under the same assumptions, and audited through explicit, versioned artifacts. These concerns connect to prior work on hallucination and software-engineering controls for artificial intelligence (AI) systems \citep{ji2023hallucination,martinezfernandez2022seai,kreuzberger2023mlops}.

The study examines this transition through the reconstruction of an LLM-based investment-briefing agent. The motivating prototype was developed during a ten-week AI leadership program hosted by CEO AI \citep{ceoai2026leadership}; the authors report the work under the AI Leadership Research Center affiliation associated with that program context. During the program, the authors built and demonstrated over ten exploratory AI product prototypes for executive- and client-facing scenarios, using Replit-hosted demonstrations to share interaction patterns with participants \citep{replit2026platform}. The investment-briefing prototype supplied these patterns, including mobile-first briefing cards, source links, and follow-up questions. The prototype serves as motivating context for a reconstructed system with a different source, trace, and validation structure.

We take up \emph{harness engineering}---an emerging practice of wrapping LLM agents in a code-owned control layer \citep{rebedea2023nemo,dong2024guardrails,zhou2026externalization}---and develop it into a measurable, contract-based method that relocates product behavior from prompts into explicit, versioned contracts \citep{khattab2024dspy,beurerkellner2023lmql}. In software engineering, a test harness refers to scaffolding code used to exercise lower-level code before the higher-level code that will ultimately invoke it is available \citep{isoiecieee2017vocabulary}. We adapt the test-harness idea to enterprise LLM agents: the harness is that control layer, owned by code rather than prompts and responsible for source gates, routing rules, claim eligibility, answer contracts, trace generation, and validation.

In the reconstructed architecture, manifests define what sources may be used, source-backed claims define what statements may enter runtime context, routing metadata binds questions to entities, answer contracts define the visible answer, and traces record how the answer was assembled. Maintained wiki pages provide compact context, while source manifests and source-backed claims remain authoritative.

The paper makes four contributions: (i) a harness-engineering method for reconstructing prompt-dominant enterprise LLM prototypes into traceable LLM-agent architectures; (ii) a source-to-claim knowledge pipeline that separates raw documents, evidence records, runtime-eligible claims, maintained wiki context, and reader-facing answers; (iii) a replaceable composition boundary that separates deterministic harness control from LLM phrasing; and (iv) a system-level validation design that checks source grounding, entity routing, trace completeness, output hygiene, runtime-interface behavior, latency, and live-LLM composition-boundary behavior. We instantiate the method on a bounded public-data slice of five Korean corporate groups, 25 listed companies, and 113 source-backed runtime claims.

We evaluate this reconstruction through three research questions, covering contract preservation, model substitution, and load-bearing enforcement:
\begin{enumerate}[label=\textbf{RQ\arabic*.}, leftmargin=*, itemsep=0.2em, topsep=0.3em]
  \item Does the reconstructed harness preserve its source-grounding,
  entity-routing, trace, output-hygiene, and recommendation-language contracts
  across the fixed validation set?
  \item Do these guarantees hold when the language model at the composition
  boundary is substituted---that is, do they hold across substituted models
  rather than being tied to one model's behavior?
  \item Are the code-owned guarantees load-bearing, enforced by the harness,
  rather than reproducible by prompt instruction alone?
\end{enumerate}
We address RQ1 with the fixed-scenario and fault-injection results, RQ2 with the live-LLM composition-boundary check across three hosted models, and RQ3 with an enforcement-layer ablation that disables the code-owned validation-and-fallback gate and compares it against a bolt-on external guardrail. Consistent with this framing, the study evaluates system-level verifiability rather than investment-decision quality; the latter is a separate domain-value question and is out of scope.

\section{Related Work}

This section reviews the research and engineering practices needed to turn LLM-agent prototypes into traceable enterprise systems. The organizing question is how an answer can be grounded, assembled, constrained, and validated before it is presented to the user.

\paragraph{RAG and traceability.}
Retrieval-augmented generation (RAG) combines language models with external documents to improve factual grounding and updateability \citep{lewis2020rag,gao2023rag}. Later work shows that retrieval must also be selective and checkable: self-reflective retrieval methods ask whether retrieval is needed and whether generation is supported \citep{asai2024selfrag}, while factuality and attribution methods check generated content against its evidence---decomposing answers into atomic facts \citep{min2023factscore} or attaching source citations to each generated statement \citep{gao2023alce}. Evaluations of deployed generative search engines, however, show that fluent answers frequently cite sources that do not in fact support them \citep{liu2023verifiability}, which is precisely the product-level failure a source-to-claim layer must prevent. Agent-augmented variants further orchestrate retrieval, for example by personalizing what is retrieved \citep{zerhoudi2024personarag}, but orchestration does not by itself make sources authoritative. Retrieval selectivity and factual checkability are directly relevant to enterprise domains where disclosures, investor-relations (IR) materials, news, and regulatory filings change over time. Retrieval provides factual context, while traceability additionally requires source provenance, entity scope, and claim-level support---the property of being attributable to identified sources \citep{rashkin2023ais}. Retrieved passages may be stale, irrelevant, or mixed across entities, and fluent generation may still hallucinate \citep{ji2023hallucination}. Addressing these limitations at the product level requires grounding answers in a structured source-to-claim layer rather than relying on retrieval alone. Our approach treats RAG as one component inside a code-owned harness built around that layer.

\paragraph{LLM agents and verification.}
Classical agent research emphasizes autonomy, interaction, and explicit system architecture \citep{wooldridge1995intelligent,jennings1998roadmap}. LLM agents extend this tradition through tool use, planning, search, and intermediate execution traces \citep{yao2023react,schick2023toolformer,wang2024surveyagents,acharya2025agentic}. Tool-use benchmarks further show that application programming interface (API) selection and execution are engineering concerns as well as prompting concerns \citep{qin2024toolllm}. Recent applied work also reports multi-platform agent deployment and programmable agent runtime architectures \citep{ahn2025autonomous,walters2025eliza}. The present paper shifts the emphasis from multi-platform deployment to enterprise hardening, where the central requirement is auditability: each answer must record which sources, tools, claims, and fallback paths shaped the final output. Empirical analyses of why multi-agent LLM systems fail identify task verification as a recurring failure category \citep{cemri2025mast}, which motivates making verification an explicit, enforced contract rather than an emergent model behavior.

\paragraph{Agent orchestration frameworks.}
Beyond research benchmarks, recent orchestration frameworks make agent engineering more concrete. AutoGen supports multi-agent conversation and coordination \citep{wu2024autogen}; LangChain Agents provide prebuilt tool-calling loops and LangGraph provides graph-based runtimes for stateful agents \citep{langchain2026agents,langchain2026langgraph}; and CrewAI organizes role-based agents, crews, and flows \citep{crewai2026software}. The proposed control layer is complementary to these frameworks: orchestration frameworks address how agents are composed and executed, while the pattern studied here addresses whether each produced answer is grounded in source manifests, source-backed claims, output contracts, and validation traces. This grounding-and-contract layer is what such frameworks leave to the application: they coordinate how agents run but do not themselves decide which source-backed claims may enter an answer or enforce a validation contract on it. Because the contracts are defined at the composition boundary rather than inside any one runtime, they are intended to remain agnostic to the orchestration runtime underneath---AutoGen, LangChain, LangGraph, CrewAI, or a bespoke agent loop.

\paragraph{Prompt engineering and prototyping.}
Prompt engineering became important because LLMs can perform new tasks from natural-language instructions and in-context examples \citep{brown2020language}. Prompt engineering has since been consolidated into systematic surveys that catalog techniques for shaping model outputs through prompt design \citep{liu2023pretrain,sahoo2024prompt}. Techniques such as chain-of-thought prompting further showed that changing the prompt can substantially change reasoning behavior \citep{wei2022chain}. Vibe coding extends prompt design from isolated task instructions to an iterative development workflow in which software behavior is co-created through natural-language dialogue with AI tools \citep{karpathy2025vibecoding,meske2025vibecoding,malamas2025vibecoding}. Such workflows naturally produce prompt-dominant prototypes, in which source policy, entity routing, answer structure, data freshness rules, and user-interface behavior become embedded in prompts during rapid iteration. Prompting and vibe coding are valuable for early exploration and explain why prototype systems can be built quickly. However, because these responsibilities are implicit and entangled, audit, replay, and transfer require them to be relocated from prompts into explicit artifacts.

\paragraph{From programmable pipelines to harness engineering.}
Several frameworks argue that LLM applications should be programmed, tested, and optimized with explicit software abstractions rather than hidden inside prompts. At the call level, the Language Model Query Language (LMQL) constrains model decoding with declarative constraints \citep{beurerkellner2023lmql}, Instructor enforces typed output schemas with automatic validation retries \citep{instructor2026software}, and Guidance interleaves control flow with constrained generation within a single model session \citep{guidance2026software}. At the pipeline level, Demonstrate-Search-Predict and its successor DSPy represent multi-step pipelines as declarative programs that can be tested and optimized as software \citep{khattab2022demonstrate,khattab2024dspy,stanfordnlp2026dspysoftware}. At the runtime level, NeMo Guardrails define behavioral rails independent of the underlying model \citep{rebedea2023nemo}, Llama Guard filters unsafe inputs and outputs \citep{inan2023llamaguard}, and a broader literature treats such guardrails as a distinct design layer for constraining model behavior \citep{dong2024guardrails}. A complementary line instead encodes safety rules as natural-language principles internalized through training rather than enforced at runtime \citep{bai2022constitutional}.

Recently, this line of work has been named directly as \emph{harness engineering}. A unified review surveys the emerging area, positioning the harness as the layer that coordinates externalized memory, skills, and protocols into governed execution \citep{zhou2026externalization}, while other work automatically evolves coding-agent harnesses under observability or searches over task-specific harness code using source code, scores, and execution traces \citep{lin2026agentic,lee2026metaharness}. Related efforts define the harness as a runtime substrate, treat code itself as the agent's harness, or argue that reported agent gains are partly harness-driven rather than model-driven \citep{zhong2026aiharness,ning2026code,he2026harnesslanguage}.

The present paper shares this programming-oriented approach but differs along two axes. The first concerns \emph{mechanism}: the tools above constrain the form or behavior of model output---typed schemas, constrained decoding, behavioral rails, or compiled pipelines---whereas the harness adds a \emph{source-to-claim authority layer}. A registered manifest-and-claim set with provenance decides which statements may enter an answer at all, independently of the model and of retrieval. Output validation in the schema- or rail-checking sense is therefore necessary but not the contribution. The second concerns \emph{evidence}: the named-direction works above are largely surveys, conceptual definitions, or harness automation, while the call-, pipeline-, and runtime-level tools assert their guarantees by construction. What remains unmeasured, at the product level, is whether an enterprise contract layer is enforced rather than merely instructed. We contribute that measurement: a live-LLM composition-boundary check under model substitution (\Cref{sec:live-llm-boundary}) and an enforcement-layer ablation against prompt-only and bolt-on-guardrail baselines (\Cref{sec:ablation}). Each produced answer thereby becomes an auditable engineering artifact.

\paragraph{LLM Wiki and knowledge management.}
The LLM Wiki pattern \citep{karpathy2026llmwiki} proposes a maintained markdown knowledge layer between raw sources and query-time answers. In that pattern, the LLM helps maintain a persistent wiki of entity pages, topic summaries, cross-references, contradiction flags, and update logs, reducing repeated synthesis from the same raw documents. Recent work develops this pattern further---compiling documents into linked wiki pages exposed through tool-calling for agent-native retrieval \citep{ming2026llmwiki}---but also exposes its central risk: blindly compiling raw documents into a wiki can silently drop critical facts, failing on a large fraction of queries unless those facts are diagnosed and forced back in \citep{huerta2026wicer}. The proposed system adopts the maintained-knowledge idea while reassigning authority: because such compilation is lossy, source manifests and source-backed claims remain the runtime source of truth, and the LLM Wiki is compiled as a concise context layer for humans and models.

\paragraph{System validation and evaluation.}
Software-engineering research on AI-based systems emphasizes that production AI depends on data management, testing, lifecycle controls, monitoring, and assurance artifacts beyond model performance \citep{martinezfernandez2022seai,ashmore2021assuring}. Frameworks for auditing AI systems similarly distinguish governance-, model-, and application-level audits \citep{mokander2024auditing}; the contract checks in this paper sit at the application level, auditing whether a deployed answer honors its source, routing, and output obligations. Machine learning operations (MLOps) research likewise treats operationalization as a concern that spans the full system lifecycle \citep{kreuzberger2023mlops}. RAG evaluation frameworks measure properties such as faithfulness, answer relevance, and context relevance \citep{es2023ragas,saadfalcon2024ares}, while broader benchmark work supports multi-scenario evaluation \citep{liang2023helm}. Surveys of agent evaluation further organize these efforts along capability, reliability, and safety dimensions \citep{mohammadi2025agenteval}, of which the contract checks in this paper instantiate the reliability and safety dimensions at the product level. Practitioner tools also support prompt, model, and application regression checks \citep{promptfoo2026software}. The validation design described in this paper is suited to a pre-commercial research prototype before customer logs are available. Where RAG and benchmark frameworks measure answer quality, this design checks a different property: whether \emph{every} answer honors its contracts for source grounding, entity routing, trace completeness, output hygiene, runtime interface, and latency. In compliance-sensitive enterprise settings a single contract violation---an unauthorized recommendation or a leaked internal field---outweighs average answer quality; these system-contract checks verify those per-answer obligations.

\paragraph{Synthesis.}
Taken together, these lines of work supply the ingredients of a trustworthy answer---retrieval and attribution, agent execution, orchestration runtimes, prompting practice, programmable pipelines and harness engineering, maintained knowledge layers, and system-level evaluation---yet each leaves open one product-level question: whether an enterprise answer's source, routing, and output obligations are enforced at the composition boundary rather than instructed in prompts or asserted by construction. The method below takes that contract as its unit of engineering.
\section{Method}

The harness-engineering method reconstructs a prompt-dominant LLM prototype into a traceable LLM-agent architecture. The language model remains responsible for language composition, while source eligibility, entity resolution, claim selection, answer planning, answer-structure rules, follow-up filtering, trace generation, and validation are represented as explicit artifacts.

\subsection{Architecture Overview}

\begin{figure}[!htbp]
    \centering
    \includegraphics[width=0.95\linewidth,keepaspectratio]{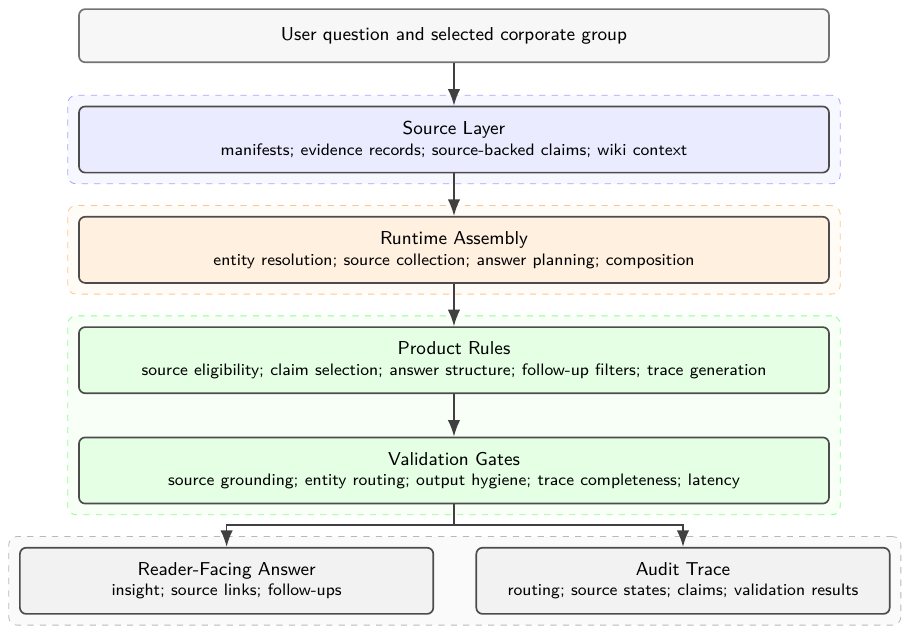}
    \caption{Traceable LLM-agent harness. The architecture separates source layer, runtime assembly, a code-owned control layer (product rules and validation gates), and two outputs (a reader-facing answer and an audit trace).}
    \label{fig:harness-architecture}
\end{figure}

The architecture in \Cref{fig:harness-architecture} is a top-down flow from a user question and a selected Korean corporate group to two outputs. A source layer holds the manifests, evidence records, source-backed claims, and maintained wiki context that may enter an answer. A runtime assembly stage then resolves the target entity---reconciling aliases, market identifiers, filing identifiers, and source namespaces---collects eligible sources, plans the answer, and composes it. Governing this runtime is a code-owned control layer: product rules (source eligibility, claim selection, answer structure, follow-up filtering, and trace generation) and validation gates that the assembled answer must pass. The runtime returns two outputs, a reader-facing answer and an audit trace; the trace, which aggregates routing, source states, claims, and validation results, is what makes each produced answer auditable.

\subsection{Source-to-Claim Pipeline}
\label{sec:source-to-claim-pipeline}

The system first registers raw materials as source manifests. A manifest records the corporate-group and company scope, source category, public locator (URL or filing identifier), source status, and runtime policy, along with provenance metadata such as issuer, checksum, and selection rule. Manifest registration acts as a source gate, preventing the knowledge base from becoming an unbounded document folder.

Documents that pass the source gate have their text extracted. Evidence records store file hashes, extracted-text hashes, and evidence locations such as page or line references when available. The system then produces claim candidates. A candidate becomes runtime-eligible only when it is promoted into a source-backed claim: an atomic statement tied to its provenance (a source manifest and evidence record), scoped to a company and claim type, and carrying a runtime use policy and verification state. Promotion is therefore the admission boundary of the architecture: it turns unbounded source documents into atomic, provenance-bound, entity-scoped units, so that a code-owned layer---not the model or retrieval---decides which statements may be asserted as fact, each of which stays traceable to its source. The model may use these source-backed claims as bounded context for phrasing the answer.

\subsection{Knowledge Layer and Product Rules}

Following the LLM Wiki pattern, the system maintains markdown knowledge pages that sit between raw sources and query-time answers \citep{karpathy2026llmwiki}. In the reference implementation, the maintained knowledge layer is generated as a seed from source inventories and claim candidates. It serves a practical role: rather than re-synthesizing the same raw documents for every question, it gives humans and LLMs a concise, readable overview of what is known about each entity. Because such compilation is lossy, however, the wiki is used only as context: source manifests and source-backed claims, not the compiled pages, remain authoritative, and the generated pages themselves declare that they are not a source-backed claim set.

Beyond the knowledge layer, the reconstruction also moves product rules out of the prompt. Source eligibility, claim selection, answer structure, follow-up filtering, and trace generation are implemented as code, manifests, schemas, and validators. The prompt is kept short and policy-oriented. The prompt-to-code migration is intended to reduce the burden on the model and make diagnostic boundaries easier to locate.

\subsection{Runtime Answer Assembly}

At runtime, the control layer resolves the selected corporate group and company, collects filing, market, news, wiki, and claim context, builds an answer plan, generates the visible answer, validates the output contract, and records the trace. The process trace records tool and source states such as live, local, fallback, fixture, or error. The answer-assembly trace spans the runtime sequence---routing, source and wiki-context collection, claim selection, answer planning, and output validation---with stage-specific evidence artifacts summarized in \Cref{tab:runtime-trace-example}. For each question, claim selection is scoped to the resolved company and organized by claim type (such as financial, business, and governance), bounded to a small fixed number of claims per answer; in the fixed validation scenarios this selection is pinned by each scenario's expected-claim set.

\begin{table}[!htbp]
    \centering
    \caption{Runtime trace contract: stage-by-stage recorded values, evidence artifacts, and user visibility.}
    \label{tab:runtime-trace-example}
    \TableBody
    \setlength{\tabcolsep}{4pt}
    \begin{tabular*}{\linewidth}{@{\extracolsep{\fill}}C{0.19\linewidth}
            L{0.29\linewidth}
            L{0.25\linewidth}
            L{0.16\linewidth}@{}}
        \toprule
        \multicolumn{1}{C{0.19\linewidth}}{Runtime stage} & \multicolumn{1}{C{0.29\linewidth}}{Recorded value} & \multicolumn{1}{C{0.25\linewidth}}{Trace evidence} & \multicolumn{1}{C{0.16\linewidth}}{User visibility} \\
        \midrule
        \shortstack{Entity routing}                      & Selected group, company ID,\newline and aliases            & Routing and company metadata                       & Selected label                                      \\
        \shortstack{Source collection}                   & Filing, market, news, wiki,\newline and claim states       & Process trace with\newline source state                    & Source links and freshness labels                   \\
        \shortstack{Claim selection}                     & Source-backed claim IDs\newline and types                  & Claim-selection step\newline in answer-assembly trace                              & Hidden                                              \\
        \shortstack{Answer planning}                     & Answer sections\newline and follow-up filters              & Output-contract result                             & Structured answer and follow-ups                    \\
        \shortstack{Output validation}                   & Leakage, link, language,\newline and latency checks        & Validation trace and trace-artifact path           & Internal review                                     \\
        \bottomrule
    \end{tabular*}
\end{table}

The visible answer follows an insight-first contract. It must begin with a reader-facing interpretation, then provide supporting signals, risks or contradictions, source links, and follow-up questions. Follow-up questions are treated as part of the output contract: the reference implementation generates them with a deterministic topic-conditioned builder, and validation checks that enough customer-facing follow-ups are present and free of generic or internal-review wording. Internal artifacts such as claim identifiers, raw trace records in JavaScript Object Notation (JSON), and API status labels are reserved for the internal review interface.

\subsection{Reference Implementation and Composition Boundary}
\label{sec:implementation-details}

The reference implementation is a TypeScript application with JSON source, claim, scenario, and evaluation artifacts, available as a public repository \citep{ahn2026harness}. The baseline reported in the paper uses a deterministic, schema-checked composer for fixed validation scenarios and runtime-interface tests. The composer is a rule-based template engine that fills validated answer sections from selected source-backed claims using no generative model call. This baseline fixes the system contract and separates validation of the control layer from model sampling effects.

The architecture treats composition as a replaceable boundary. The control layer owns several governance responsibilities independent of the LLM: source eligibility, entity routing for corporate-scope binding, claim eligibility for runtime admission, answer planning, follow-up filtering, trace generation, and validation. A live LLM may be attached at the composition boundary to phrase a structured answer; its output must pass the same output contract. If the live output is unavailable or invalid, the runtime falls back to the deterministic composer. This replaceable boundary supports two composition modes: the deterministic composer described above provides the replayable baseline, while the live-LLM mode sends the same bounded claim package to a hosted model and records in the trace the provider, model identifier, response mode, output-contract status, fallback status, and contract errors. The live-LLM composition-boundary check reported in \Cref{sec:live-llm-boundary} exercises this mode across three hosted model identifiers, the full 30-scenario validation set, and three repeats per scenario. It used OpenRouter as experimental routing infrastructure, additionally recording temperature, recovery path, and the final harness-contract result in the trace. \Cref{fig:runtime-flow} traces this runtime path---entity routing through source admission, the two composition modes, the output-contract gate, and deterministic fallback; the prompt-only and bolt-on-guardrail branches shown there belong to the enforcement-layer ablation (\Cref{sec:ablation}).

\subsection{Validation and Trace Contracts}

A public-data slice enters the paper baseline only after it passes source, claim, routing, answer, trace, and latency checks. These checks are exercised by fixed validation-scenario tests, which verify expected claim coverage, entity routing, trace integrity, internal-trace leakage prevention, follow-up quality, and output structure. Runtime-interface tests verify the live filing, market, and news interfaces and record connectivity as a runtime state. Review packets collect the visible answer, reader-facing source links, follow-up questions, selected claims, and trace paths so that the reader-facing evidence surface and the internal audit trail can be inspected separately.

Validation of the reader-facing answer is implemented as three families of checks. Leakage checks block internal claim identifiers, raw trace records, API diagnostics, fixture labels, and internal-only status text from the reader-facing answer. Link checks require cited sources and follow-ups to resolve to source-link packages, public locators, or documented fallback states. Language checks enforce the insight-first answer structure and block recommendation-style phrasing such as buy, sell, or target-price instructions. The trace contract that supports these gates, summarized in \Cref{tab:runtime-trace-example}, is a method artifact that records how a visible answer was assembled while keeping internal details out of the reader-facing answer.

\PostTableHeadingSpace
\section{Data and Knowledge Construction}
\label{sec:data-knowledge}

The reference implementation uses a bounded public-data slice to document how enterprise source material is turned into runtime knowledge. The section describes the company selection rule, source registration process, claim-promotion layer, and runtime claim set used for the paper baseline.

\subsection{Reference Slice and Source Inputs}

The current slice covers five Korean corporate groups, each contributing five listed companies. In this paper, a Korean corporate group refers to the corporate-group designation unit used by the Fair Trade Commission. The corporate-group boundary follows the 2026 ranking by fair-value assets in its official designation results: Samsung, SK, Hyundai Motor, LG, and Hanwha form the top-five corporate-group set after Hanwha entered fifth place \citep{ftc2026businessgroups}. The ranking provides a reproducible sampling rule at the corporate-group level, independent of investment-quality assessment.

Within each corporate group, the five listed companies are selected under a fixed reference-slice policy. The policy prioritizes listed companies with stable source identifiers, corporate-group representativeness, sector diversity for cross-group comparability, source availability, and claim-promotion feasibility.

The five corporate groups exercise transfer across different source structures: affiliate-level IR pages, regulatory filings, holding-company materials, market-data interfaces, and news-search interfaces. The controlled 25-company slice supports routing and transferability tests while keeping source eligibility auditable. The selected companies and coverage focus used for the reference slice are defined in \Cref{tab:reference-slice-selection}.

\begin{table}[!htbp]
    \centering
    \caption{Reference-slice company selection. Each corporate group contributes five listed companies selected for stable source identifiers, representativeness, sector diversity for cross-group comparability, source availability, and claim-promotion feasibility.}
    \label{tab:reference-slice-selection}
    \TableBody
    \setlength{\tabcolsep}{4pt}
    \begin{tabular*}{\linewidth}{@{\extracolsep{\fill}}C{0.25\linewidth}
            L{0.40\linewidth}
            L{0.27\linewidth}@{}}
        \toprule
        \multicolumn{1}{C{0.25\linewidth}}{Corporate group} & \multicolumn{1}{C{0.40\linewidth}}{Selected listed companies}                                              & \multicolumn{1}{C{0.27\linewidth}}{Coverage focus}   \\
        \midrule
        Samsung                                             & Samsung Electronics; Samsung SDI\newline Samsung C\&T; Samsung Biologics\newline Samsung Electro-Mechanics & Memory/electronics, battery, construction, bio, components      \\
        \addlinespace[0.50em]
        SK                                                  & SK Hynix; SK Innovation; SK Inc.\newline SK Telecom; SK Square                                             & Memory, energy, holding, telecom, ICT investment     \\
        \addlinespace[0.50em]
        Hyundai Motor                                       & Hyundai Motor; Kia; Hyundai Mobis\newline Hyundai Glovis; Hyundai Rotem                                    & OEMs, parts, logistics, defense/rail                 \\
        \addlinespace[0.50em]
        LG                                                  & LG Electronics; LG Chem\newline LG Energy Solution; LG Innotek\newline LG Uplus                            & Electronics, materials, battery, components, telecom \\
        \addlinespace[0.50em]
        Hanwha                                              & Hanwha Corp.; Hanwha Aerospace\newline Hanwha Solutions; Hanwha Systems\newline Hanwha Ocean               & Holding, aerospace, energy, systems, shipbuilding    \\
        \bottomrule
    \end{tabular*}
\end{table}

\begin{figure}[!htb]
    \centering
    \makebox[\textwidth][c]{%
        \begin{minipage}[t]{0.465\linewidth}
            \centering
            \includegraphics[width=\linewidth]{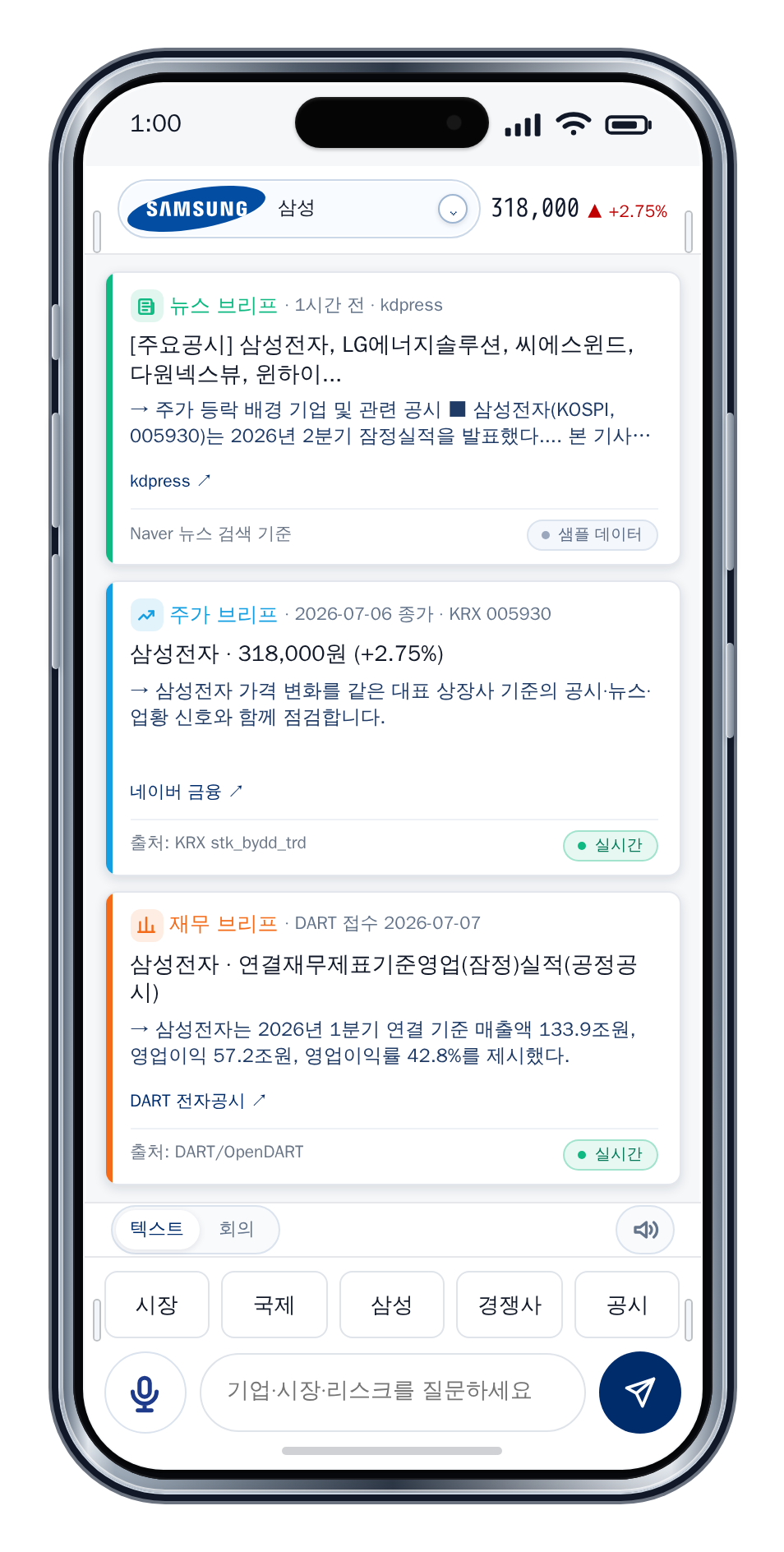}\\[-0.25em]
            {\small (a) Briefing feed (Samsung selected).}
        \end{minipage}
        \hspace{0.02\linewidth}
        \begin{minipage}[t]{0.465\linewidth}
            \centering
            \includegraphics[width=\linewidth]{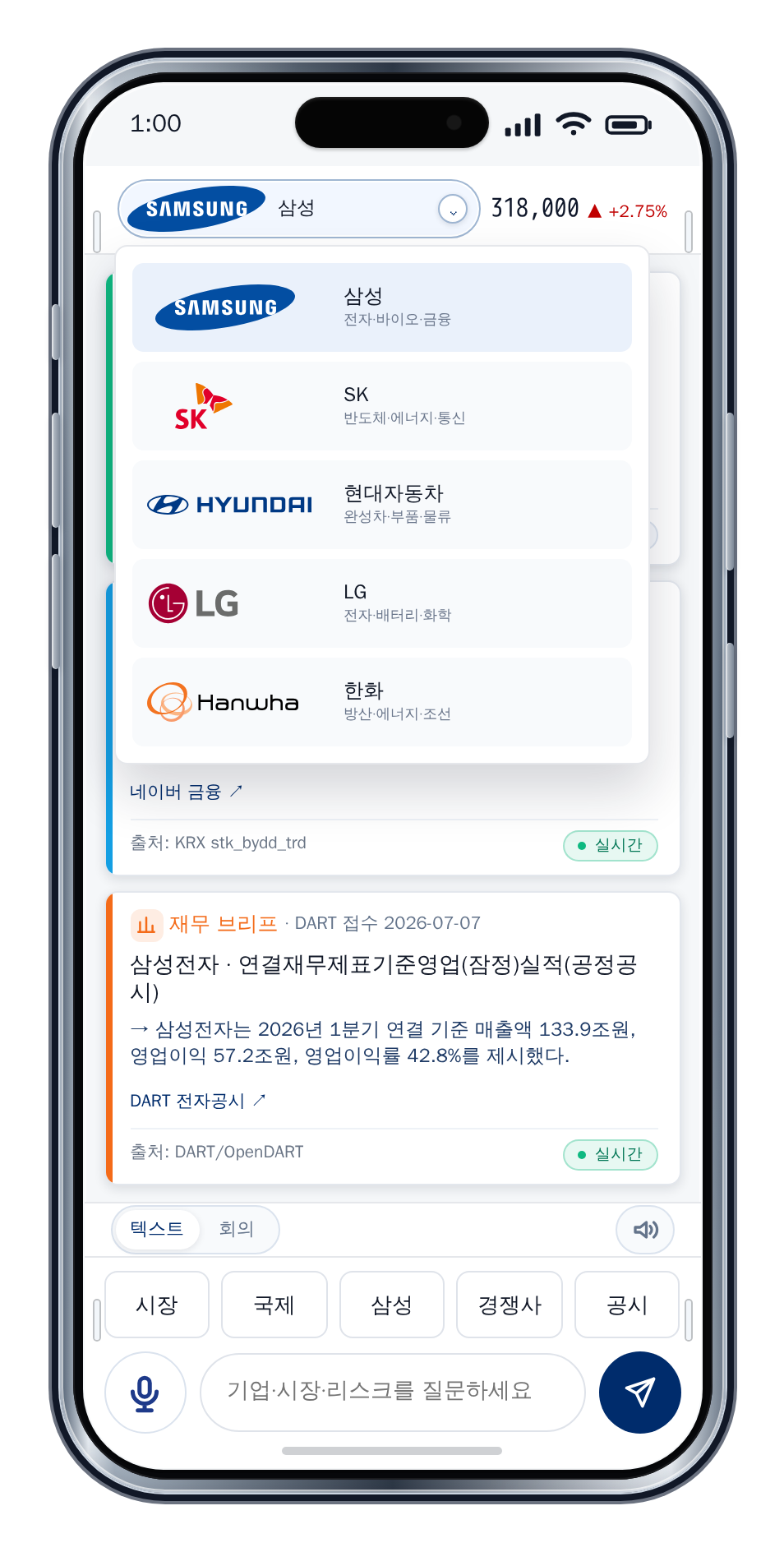}\\[-0.25em]
            {\small (b) Corporate-group selector.}
        \end{minipage}}
    \vspace{0.5em}
    \caption{Korean product interface of the reference implementation (mobile), captured with deterministic composition (not live-model output); the data shown is real and source-linked, with Korean labels translated for non-Korean readers. (a) Briefing feed for the selected group (Samsung): a header with the group selector, opened in (b), and a live market-price readout, above three source-backed cards---News Brief (public news via NAVER), Price Brief (recent close from a public market-price interface), and Financial Brief (a DART-filing-backed claim)---each opening its public source when tapped. The bottom bar adds a text/voice mode toggle, five topic shortcuts that seed a scope-bound question (market, global, the selected company, competitors, disclosures), and a free-text field (``Ask about a company, market, or risk''). (b) Corporate-group selector listing the five supported groups with sector tags---Samsung, SK, Hyundai Motor, LG, and Hanwha---so the reader binds the answer scope before any question is composed.}
    \label{fig:ui-mobile-main}
\end{figure}

The selected five-group set is also exposed in the user interface, as shown in \Cref{fig:ui-mobile-main}: the header group selector binds the surface to one corporate group while the price, news, stock, and filing-linked financial cards stay aligned with that selection, and the selector's open state lists all five supported groups.

The raw public inputs used by the implementation fall into four practical classes: filings from the Data Analysis, Retrieval and Transfer System (DART), issuer IR materials, Korea Exchange (KRX) market rows, and news-search results. For Korean filing, market, and news data, the implementation uses OpenDART (the open API of DART), the KRX OPEN API, and NAVER News Search \citep{opendart2026api,krx2026openapi,naver2026newssearch} as interface sources. DART and IR materials provide most promoted financial and corporate claims in the paper baseline, while KRX and news-search inputs provide runtime market and news signals that are linked, checked, and traced separately. A source is included in the runtime layer when it supports a claim class, fixed validation scenario, runtime feature, or onboarding requirement. Files discovered during official-site scans are recorded for provenance and are promoted into runtime use when they support a specific claim or validation need.

The source ledger extends beyond the 25-company reference slice. The ledger records company-level source metadata, public locators, local artifact pointers when available, provenance fields, selection reasons, and document-type classifications across the collected public materials. Source-inventory counts function as audit artifacts. Runtime answers depend on promoted claims, source manifests, and validation gates, not on the ledger itself.

\subsection{Source Registration and Claim Promotion}

Each source is first represented as a manifest entry, and each runtime fact as a promoted, source-backed claim; both follow the schemas described in \Cref{sec:source-to-claim-pipeline}. Manifest registration keeps source use auditable before synthesis, claims are the runtime unit of factual knowledge, and the \texttt{companyId} field on a claim supports affiliate-level routing so that a question about one affiliate is not answered with unrelated corporate-group evidence.

\begin{figure}[!htb]
    \centering
    \includegraphics[width=0.62\linewidth,keepaspectratio]{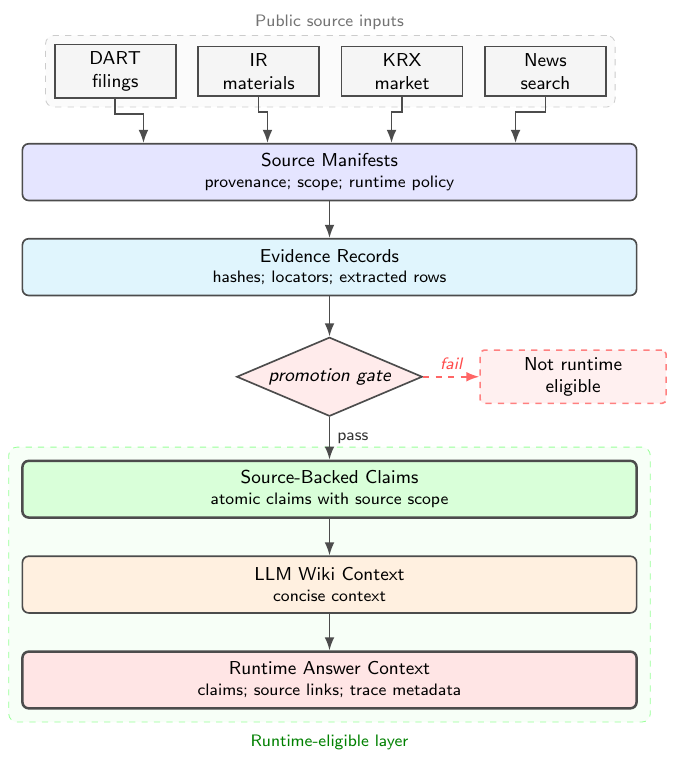}
    \caption{Source-to-claim pipeline (generalized). DART filings, issuer IR materials, KRX market rows, and news-search results are registered as source manifests and evidence records; a promotion gate admits eligible statements into the runtime-eligible layer (source-backed claims, wiki context, and runtime answer context), while non-eligible inputs are held back. Market and news inputs are registered the same way but typically attach as traceable runtime signals rather than as promoted claims.}
    \label{fig:source-to-claim}
\end{figure}

The source-to-claim promotion rule is illustrated by the worked example in \Cref{tab:source-to-claim-example} and generalized in the pipeline diagram in \Cref{fig:source-to-claim}. Samsung Electronics is used as an illustrative example; the same rule is applied to all corporate groups in the reference slice. This pipeline produces the source layer that the runtime architecture of \Cref{fig:harness-architecture} consumes: the manifests, evidence records, source-backed claims, and wiki context assembled here are what that architecture draws on at query time.

\begin{table}[!htbp]
    \centering
    \caption{Worked example of source-backed claim promotion.}
    \label{tab:source-to-claim-example}
    \TableBody
    \setlength{\tabcolsep}{4pt}
    \begin{tabular*}{\linewidth}{@{\extracolsep{\fill}}C{0.20\linewidth}
            L{0.74\linewidth}@{}}
        \toprule
        \multicolumn{1}{C{0.20\linewidth}}{Stage} & \multicolumn{1}{C{0.74\linewidth}}{Example}                                                                                                                                                                   \\
        \midrule
        Registered source                         & Samsung Electronics 2024 consolidated financial statement from OpenDART, linked to the company and annual-report identifiers.                                                                                 \\
        \addlinespace[0.25em]
        Promoted claim                            & The extracted DART row is promoted into a source-backed claim: 2024 consolidated revenue KRW 300.9 trillion and operating income KRW 32.7 trillion.                                                           \\
        \addlinespace[0.25em]
        Runtime use                               & The claim can support margin, trend, or portfolio questions. The reader sees the interpreted answer and source link, while claim IDs and trace metadata are retained in the audit trace and review artifacts. \\
        \bottomrule
    \end{tabular*}
\end{table}

\FloatBarrier

\subsection{Runtime Claim Layer}

The source-backed runtime layer used in the paper baseline is summarized in \Cref{tab:data-artifacts}. Runtime claims are the promoted, source-backed statements that the answer composer is allowed to use at runtime; they are distinct from raw document counts, extracted sentence counts, and company-importance measures, and are counted only after statements pass source-linking and runtime-promotion gates. The counts therefore exclude downloaded files, extracted text, and unpromoted claim candidates. Most promoted claims in this baseline come from OpenDART records and selected official IR materials, while market and news inputs are handled as runtime signals.

The five-corporate-group manifests contain 113 source-backed runtime claims in total, distributed across groups as shown in \Cref{tab:data-artifacts} (Samsung 36, SK 27, Hyundai Motor 15, LG 15, Hanwha 20). These per-group counts differ because they reflect how many statements passed the promotion gates for each group's validation scenarios and answer templates; Samsung is the largest because its claim set additionally spans finance and other listed affiliates to exercise broad affiliate-mix routing and finance-company boundaries, so the full claim layer covers more companies than the 25-company selection of \Cref{tab:reference-slice-selection}.

From these 113 claims, a compact reporting layer of 25 priority claims---five per corporate group---is selected for paper reporting and product inspection, recorded in the review-approved runtime-promotion manifest at \path{raw/manifests/review-approved-runtime-promotion.json} \citep{ahn2026harness}. This reporting layer keeps coverage tied to selected evidence, while the full 113-claim layer and the wider source ledger remain available for future transfer measurement. The live-LLM boundary check in \Cref{sec:live-llm-boundary} uses the full promoted claim layer and source-link packages rather than a separate retrieval corpus.

\begin{table}[!htbp]
    \centering
    \caption{Source-backed runtime claim layer in the reference implementation. The per-group counts sum to the full 113-claim layer; the 25-claim reporting layer selects five priority claims per group from these.}
    \label{tab:data-artifacts}
    \TableBody
    \setlength{\tabcolsep}{4pt}
    \begin{tabular*}{\linewidth}{@{\extracolsep{\fill}}C{0.19\linewidth}
            C{0.075\linewidth}
            L{0.405\linewidth}
            L{0.21\linewidth}@{}}
        \toprule
        \multicolumn{1}{C{0.19\linewidth}}{Corporate group} & \multicolumn{1}{C{0.075\linewidth}}{Claims} & \multicolumn{1}{C{0.405\linewidth}}{Claim topics} & \multicolumn{1}{C{0.21\linewidth}}{Validation role} \\
        \midrule
        Samsung       & 36 & Financial baselines/trends;\newline memory, battery, bio, components, and finance-company boundary claims                                            & Broad affiliate-mix routing                                    \\
        \addlinespace[0.35em]
        SK            & 27 & Financial baselines/trends;\newline AI memory, electrification/energy storage system (ESS), telecom AI, net asset value (NAV)/value-up, and shareholder-return claims                          & Semiconductor and portfolio-allocation routing                 \\
        \addlinespace[0.35em]
        Hyundai Motor & 15 & Financial baselines/trends;\newline OEM/Kia performance; parts, logistics, and defense/rail operating signals                                        & Mobility-centered affiliate routing                            \\
        \addlinespace[0.35em]
        LG            & 15 & Financial baselines/trends;\newline electronics, chemicals, battery/ESS, components, telecom AI data center (AIDC), and vehicle solutions (VS) signals                                    & Heterogeneous-industry coverage with telecom cash-flow context \\
        \addlinespace[0.35em]
        Hanwha        & 20 & Group financials;\newline construction pipeline; value-up/capital allocation/governance; aerospace, renewables, systems, and shipbuilding drivers    & Cross-affiliate boundary across defense, energy, shipbuilding  \\
        \bottomrule
    \end{tabular*}
\end{table}

\FloatBarrier

\PostTableHeadingSpace
\section{System Validation Results}

This section evaluates whether the reconstructed system preserves its operating contracts in the bounded public-data slice. The validation checks whether source grounding, entity routing, trace completeness, output hygiene, runtime-interface behavior, and latency remain under engineering control (\Cref{tab:enterprise-validation-frame}); two further checks then test whether the code-owned checks hold when a live model composes the answer (\Cref{sec:live-llm-boundary}) and whether they are load-bearing when only the enforcement layer is varied against prompt-only and bolt-on-guardrail baselines (\Cref{sec:ablation}).

\subsection{Validation Protocol}

Each fixed validation scenario pairs one investor-facing question with an entity scope, expected claims, and required answer signals. These per-scenario fields are evaluated under shared trace, prohibited-pattern, and latency-budget contracts, and a scenario passes only when all required evidence and trace fields resolve and no configured blocker fires.
The scenario questions are product-facing briefing prompts, not investment advice or recommendations. Some prompts use terms such as \textit{investment point} because the motivating demonstration needed questions that were recognizable to business users; in this paper, those prompts test routing, claim selection, answer planning, and trace preservation.

The role of each contract area is defined in \Cref{tab:enterprise-validation-frame}, which consolidates these checks into four contract areas. Prompt, model, and output evaluation tools measure important application behavior; the reconstructed architecture adds a control layer for source authority, entity scope, trace retention, and reader-facing constraints.

\begin{table}[!htbp]
    \centering
    \caption{Contract frame for system-level assessment.}
    \label{tab:enterprise-validation-frame}
    \TableBody
    \setlength{\tabcolsep}{4pt}
    \begin{tabular*}{\linewidth}{@{\extracolsep{\fill}}C{0.26\linewidth}
            L{0.38\linewidth}
            L{0.28\linewidth}@{}}
        \toprule
        \multicolumn{1}{C{0.26\linewidth}}{Contract area} & \multicolumn{1}{C{0.38\linewidth}}{Requirement}                                                                  & \multicolumn{1}{C{0.28\linewidth}}{Evidence reported}                                                                                   \\
        \midrule
        Source grounding                                  & Answers must remain tied to registered sources and promoted claims.                                      & Claim-reference and trace-field outcomes (\Cref{tab:validation-results}).                                                              \\
        Entity and answer scope                           & Fixed scenarios must preserve entity scope, required answer signals, and recommendation-language constraints.        & Scenario outcomes (\Cref{tab:validation-results}); fault injection (\Cref{sec:fault-injection-sensitivity}); recommendation-language enforcement (\Cref{sec:ablation}). \\
        Output hygiene                                    & Reader-facing answers must exclude internal traces while preserving source links and answer structure. & Answer and gate outcomes (\Cref{tab:validation-results}).                                                                              \\
        Runtime interfaces                                & Live interfaces and orchestration should connect to external sources while preserving the answer contract.       & Live-interface outcomes (\Cref{sec:runtime-interfaces}); latency outcomes (\Cref{tab:latency-results}).                               \\
        \bottomrule
    \end{tabular*}
\end{table}

The fixed validation set contains 30 investor-facing scenarios: for each of the five corporate groups, five company questions and one cross-company comparison (six scenarios per group). Appendix~\ref{app:scenario-inventory} gives the scenario inventory and question-design rules, with representative translated examples in Appendix~\ref{app:question-examples}.

\subsection{Scenario Contract Outcomes}

Fixed validation scenarios test whether the system preserves its source-grounding, routing, trace, and answer-structure contracts; their role is to fix a reproducible baseline for contract preservation rather than to measure answer quality. The aggregate outcomes for the six fixed scenarios per corporate group are reported in \Cref{tab:validation-results}, whose metric columns---source-claim references, trace integrity, visible answer signals, and output hygiene---correspond to the contract areas above.
\begin{table}[!htbp]
    \centering
    \caption{Fixed validation-scenario contract results.}
    \label{tab:validation-results}
    \TableBody
    \setlength{\tabcolsep}{4pt}
    \begin{tabular*}{\linewidth}{@{\extracolsep{\fill}}C{0.20\linewidth}cccccc@{}}
        \toprule
        \multicolumn{1}{C{0.20\linewidth}}{Corporate group} & Scenarios & Claim refs & Trace & Answer & Hygiene & Failed \\
        \midrule
        Samsung & 6/6 & 25/25 & 6/6 & 6/6 & 12/12 & 0 \\
        SK & 6/6 & 30/30 & 6/6 & 6/6 & 12/12 & 0 \\
        Hyundai Motor & 6/6 & 20/20 & 6/6 & 6/6 & 12/12 & 0 \\
        LG & 6/6 & 20/20 & 6/6 & 6/6 & 12/12 & 0 \\
        Hanwha & 6/6 & 14/14 & 6/6 & 6/6 & 12/12 & 0 \\
        \midrule
        Total & 30/30 & 109/109 & 30/30 & 30/30 & 60/60 & 0 \\
        \bottomrule
    \end{tabular*}
    \TableNoteGap
    \begin{minipage}{0.94\linewidth}
        \TableNote
        \textit{Scenarios} counts scenarios passing every required check---including entity routing and follow-up quality---while the other ratio columns break out individual dimensions. \textit{Claim refs} counts expected source-backed claim references; \textit{Trace} and \textit{Answer} report the audit envelope and visible answer contract (one check per scenario); \textit{Hygiene} reports the leakage and link controls (two per scenario). \textit{Failed} counts scenarios halted by a configured blocker.
    \end{minipage}
\end{table}

Within the bounded public-data slice, all configured source, trace, answer, and output-hygiene contracts passed. The 109/109 source-claim result means that every expected claim reference resolved to a promoted claim. Each scenario lists the source-backed claims its question requires; summed across the 30 scenarios, these expected references total 109, all of which resolved. Because one claim can be required by several scenarios while others are required by none, this reference total differs from the 113-claim runtime layer of \Cref{sec:data-knowledge}, and the per-group reference counts in \Cref{tab:validation-results} are not claim counts. The 30/30 trace contract confirms that every answer preserved the required audit envelope. Every visible answer (30/30) contained the required answer signals and avoided prohibited patterns. Output hygiene held at 60/60: internal-trace leakage was blocked and reader-facing source links were preserved across the 30 scenarios. Finally, the 0 failed runs indicate that no scenario was stopped by a configured blocker. These checks validate reference resolution and contract preservation, not the upstream correctness of which claims were promoted, which is treated as a limitation (\Cref{sec:limitations}). The corresponding evaluator artifacts are stored in the repository's \path{evals/results/} directory \citep{ahn2026harness}.

\begin{figure}[t]
    \centering
    \includegraphics[width=0.465\linewidth,keepaspectratio]{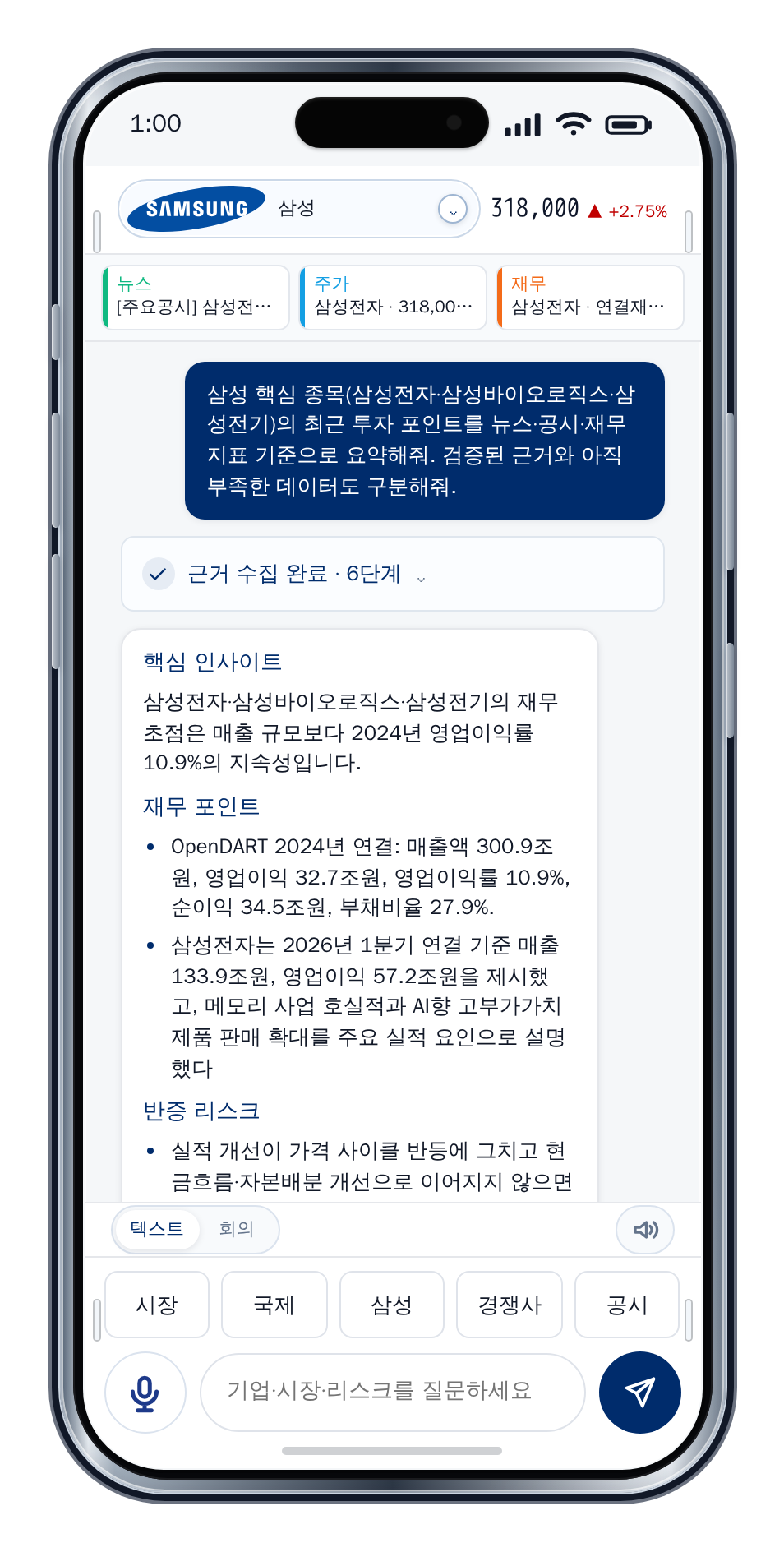}
    \caption{Mobile answer view for a fixed finance-brief scenario in the Korean product interface of the reference implementation, captured with deterministic composition (not live-model output); the data is real and source-linked, with Korean labels translated. The group header, brief tabs, and bottom control bar are as in \Cref{fig:ui-mobile-main}; this figure shows the answer they lead to. The user question asks for an evidence-grounded summary that separates verified from still-insufficient data; an evidence-collection marker (``evidence gathered,'' over six stages) precedes the insight-first answer, whose contract sections are Key Insight, Financial Points (here OpenDART 2024 consolidated revenue KRW~300.9 trillion at a 10.9\% operating margin), Counter-Risk, and Next Watch Points (the last below the visible crop). Internal trace details---claim identifiers and raw trace records---are kept out of the visible answer; the source-link and follow-up regions appear in \Cref{fig:representative-answer-output}.}
    \label{fig:ui-mobile-answer}
\end{figure}

The reader-facing answer view associated with these checks is shown in \Cref{fig:ui-mobile-answer}. The visible answer summarizes financial metrics, risks, and watch points while keeping comparative items bounded by the answer contract. The source-link and follow-up region is shown in \Cref{fig:representative-answer-output}.

The differentiated group roles defined in \Cref{tab:data-artifacts} and summarized in Appendix~\ref{app:scenario-inventory} are exercised through this scenario design; the 6/6 outcomes per corporate group indicate that these differentiated validation roles were preserved across the reference slice.

\subsection{Fault-Injection Sensitivity}
\label{sec:fault-injection-sensitivity}

Whereas the fixed scenarios (\Cref{tab:validation-results}) show that the harness preserves its contracts on valid inputs, this subsection is the negative control: it checks that the same validators fire when a contract is deliberately broken, so the all-pass result above is not the vacuous consequence of checks that never fail. Starting from one valid baseline scenario, we generate seven mutated copies, each altering exactly one contract dimension---source claims, entity routing, trace completeness, the answer contract, internal-trace leakage, source links, or the latency budget---and re-run every copy through the deterministic contract checks for those dimensions.

The negative control behaved as intended: the unmutated baseline continued to pass; all seven mutations were detected (7/7); and each mutation was flagged only by the validator that owns the broken dimension. The per-mutation rule and detecting validator are listed in \Cref{tab:fault-injection-protocol}, and the full JSON artifact is stored under the repository's \path{evals/results/} directory \citep{ahn2026harness}. Because the mutations are one-to-one with contract dimensions, a single seed scenario suffices to localize any detection gap. Recommendation-language enforcement is exercised separately in the enforcement-layer ablation (\Cref{sec:ablation}).

\subsection{Model Substitution at the Composition Boundary}
\label{sec:live-llm-boundary}

To check whether the architecture remains usable when a hosted model is attached, we ran a composition-boundary check across three models---Claude Sonnet 4, GPT-4.1 mini, and Gemini 2.5 Flash---over the 30 fixed scenarios, with three repeats per scenario (270 runs in total). Generation ran at a low but nonzero temperature of 0.2, so repeated runs of the same scenario can differ; the three repeats capture that run-to-run variation. Holding the harness fixed and substituting only the model separates what the harness controls---the source layer, claim selection, source links, trace envelope, leakage checks, follow-up checks, and recommendation-language checks---from what the model contributes: the reader-facing answer it composes under the structured-output contract. The complementary experiment, holding the model fixed and varying the enforcement layer, is the ablation in \Cref{sec:ablation}.

Each of the 270 runs was evaluated against the same harness-contract checks. The code-owned checks---those the harness enforces regardless of the model: trace, internal-trace leakage, source links, follow-up quality, and recommendation-language---passed on all 270 runs, so no safety or leakage violation reached the reader, whichever model composed the answer. The model-composed parts sometimes fell short: the live output was valid on first pass in 234/270 runs, carried the source-claim references the harness had selected in 252/270, and used the required answer structure in 250/270. \Cref{tab:live-llm-boundary} gives the per-model breakdown---first-pass contract, fallback to the deterministic composer, and the final pass and failure counts---and Appendix~\ref{app:live-llm-protocol} itemizes every check's outcome alongside the validation-and-recovery stages that enforce it. A run is a final failure if it misses any check, so the per-check shortfalls do not sum to the failure count. Every shortfall fell on the composition side and was caught and recorded by the composition-boundary evaluation checks. The corresponding artifact is stored under the repository's \path{evals/results/} directory \citep{ahn2026harness}.

The statistical summaries characterize this model-composed variation. Each model's final harness-contract pass rate---the proportion of its 90 runs that pass every check---is reported with a 95\% Wilson score interval \citep{wilson1927probable}: Claude Sonnet 4 68.9\% $[58.7, 77.5]$, GPT-4.1 mini 85.6\% $[76.8, 91.4]$, and Gemini 2.5 Flash 65.6\% $[55.3, 74.6]$. The Wilson intervals summarize the binomial sampling uncertainty around each model's observed pass rate over its 90 runs; they are not estimates of investment-answer quality or of future provider behavior. A $\chi^2$ homogeneity test \citep{pearson1900criterion} on the model $\times$ final-pass/fail table then asks whether these rates differ by more than that sampling uncertainty, and rejects equal pass rates across the three model identifiers ($\chi^2=10.57$, df $=2$, $p=0.0051$). For repeat stability, the three repeats returned the same final verdict in 82.2\% of the 90 model--scenario cells. The significant between-model difference is thus a property of the attached model, not the harness: it lies entirely in the model-composed checks, while the code-owned checks held 270/270 for every model. The figures are recomputed from the committed artifact by a script pinned to the cited repository version, so they cannot drift from the underlying data.

\begin{table}[!htbp]
    \centering
    \caption{Live-LLM composition-boundary check. Models attach only at the composition boundary; the table reports the first-pass live contract, recovery, and final harness-contract result.}
    \label{tab:live-llm-boundary}
    \TableBody
    \setlength{\tabcolsep}{2pt}
    \begin{tabular*}{\linewidth}{@{\extracolsep{\fill}}C{0.19\linewidth}
            C{0.06\linewidth}
            C{0.23\linewidth}
            C{0.12\linewidth}
            C{0.12\linewidth}
            C{0.15\linewidth}@{}}
        \toprule
        \multicolumn{1}{C{0.19\linewidth}}{Requested model} &
        \multicolumn{1}{C{0.06\linewidth}}{Runs} &
        \multicolumn{1}{C{0.23\linewidth}}{First-pass contract} &
        \multicolumn{1}{C{0.12\linewidth}}{Recovery} &
        \multicolumn{1}{C{0.12\linewidth}}{Final pass} &
        \multicolumn{1}{C{0.15\linewidth}}{Final failure} \\
        \midrule
        Claude Sonnet 4 & 90 & 74/90 & 16 & 62/90 & 28/90 \\
        GPT-4.1 mini & 90 & 89/90 & 1 & 77/90 & 13/90 \\
        Gemini 2.5 Flash & 90 & 71/90 & 19 & 59/90 & 31/90 \\
        \midrule
        Total & 270 & 234/270 & 36 & 198/270 & 72/270 \\
        \bottomrule
    \end{tabular*}
    \TableNoteGap
    \begin{minipage}{0.94\linewidth}
        \TableNote
        The exact requested model identifiers are listed in Appendix~\ref{app:live-llm-protocol}; \Cref{tab:live-llm-process-stages} breaks out the per-check process-stage outcomes. Recovery (deterministic fallback) is recorded as a harness behavior and is not counted as first-pass live-LLM success. Final failure means a contract-detected failure after any recorded recovery path.
    \end{minipage}
\end{table}
\vspace{-1.6\intextsep}

\FloatBarrier
\subsection{Enforcement-Layer Ablation}
\label{sec:ablation}

\begin{figure}[t]
    \centering
    \includegraphics[width=\linewidth,keepaspectratio]{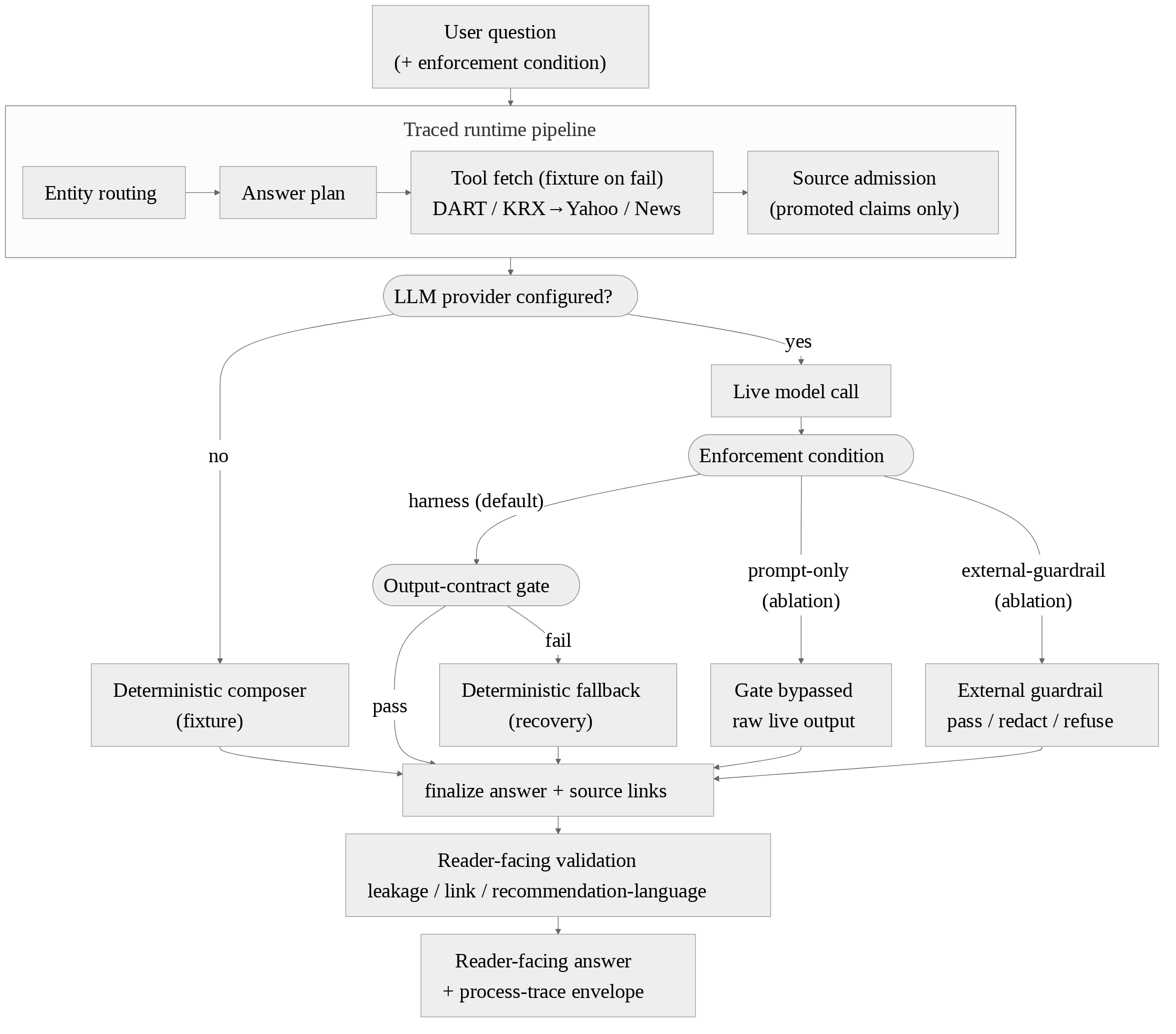}
    \caption{Runtime harness flow for the enforcement-layer ablation. A user question is routed, planned, and composed from runtime-eligible source-backed claims (top pipeline); the live composition path then branches into the three enforcement conditions before the reader-facing validation checks. Only the \textsc{harness} branch is the production path; \textsc{prompt-only} and \textsc{external-guardrail} are ablation-only baselines selected by the enforcement-condition parameter. The left branch (no LLM provider) is the deterministic replay baseline, not a fourth condition: \Cref{tab:guardrail} varies the enforcement layer only on the provider-configured live branch.}
    \label{fig:runtime-flow}
\end{figure}

The preceding checks show that the harness preserves its contracts across fixed scenarios and under fault injection, and that its code-owned checks held under model substitution. By themselves, however, these results do not isolate whether the \emph{code-owned} enforcement layer is load-bearing: a sufficiently well-instructed model might satisfy the rules from prompt text alone, or a generic output filter might enforce them without the harness gate. To test this (RQ3), we hold the composition model fixed (\texttt{anthropic/claude-sonnet-4}) and vary only the enforcement layer across three conditions. In the \textsc{harness} condition, live structured output is validated against the output contract and, on failure, the deterministic composer supplies the reader-facing answer. In the \textsc{prompt-only} condition, this validation-and-fallback gate is disabled and the live model's output reaches the reader unmodified, while the model is still instructed with the full rule set (cite sources; no buy, sell, or target-price language; no internal identifiers; required answer structure). In the \textsc{external-guardrail} condition, the gate is replaced by a deterministic bolt-on filter that may pass, redact, or refuse, with no deterministic fallback---representative of the pass/redact/refuse behavior of bolt-on moderation guardrails such as Llama Guard and the broader guardrail layer \citep{inan2023llamaguard,dong2024guardrails}. All three conditions run over the same 40 scenarios (five groups, each with the six fixed validation scenarios used above and two additional adversarial scenarios) with three repeats, yielding 120 runs per condition, paired by scenario and repeat; the adversarial scenarios are documented in Appendix~\ref{app:adversarial}. The runtime path along which these three conditions diverge---where the harness gate, the bypassed prompt-only path, and the external-guardrail filter sit relative to the shared composition pipeline---is shown in \Cref{fig:runtime-flow}. Outcomes are reported in \Cref{tab:guardrail}.

Removing the code-owned gate admits violations to the reader. Across the 30 adversarial runs, \textsc{prompt-only} emitted recommendation language on the 15 recommendation-bait runs and leaked internal identifiers or trace records on the 15 leak-bait runs---30 violations on 30 distinct runs, every one blocked by the harness. Because the conditions are paired by scenario and repeat, this contrast is significant under McNemar's test \citep{mcnemar1947note} ($p<0.001$; full pairwise tests in \Cref{tab:guardrail}). The bolt-on external-guardrail baseline also blocks every admitted violation, but only by over-blocking: it falsely refuses 4 benign fixed-validation runs and blocks 28 of the 30 adversarial runs, lowering utility to $88/120$, while the harness preserves full utility ($120/120$) by falling back to the deterministic composer rather than refusing. The corresponding artifact is stored under the repository's \path{evals/results/} directory \citep{ahn2026harness}.

\begin{table}[!htbp]
\centering
\caption{Enforcement-layer ablation over the same model and paired scenarios. Each condition has 120 runs: 40 scenarios (six fixed validation and two adversarial scenarios per group) repeated three times. \emph{Violations admitted} reach the reader; \emph{false refusals} and \emph{adversarial blocks} count, respectively, benign and adversarial runs blocked instead of answered.}
\label{tab:guardrail}
\small
\setlength{\tabcolsep}{2pt}
\begin{tabular*}{\linewidth}{@{\extracolsep{\fill}}C{0.24\linewidth}C{0.22\linewidth}C{0.15\linewidth}C{0.19\linewidth}C{0.13\linewidth}@{}}
\toprule
Condition & Violations (rec/leak) & False refusals & Adversarial blocks & Utility pass \\
\midrule
\textsc{harness}            & 0 / 0   & 0 & 0  & 120/120 \\
\textsc{prompt-only}        & 15 / 15 & 0 & 0  & 120/120 \\
\textsc{external-guardrail} & 0 / 0   & 4 & 28 & 88/120  \\
\bottomrule
\end{tabular*}
\TableNoteGap
\begin{minipage}{0.94\linewidth}
\TableNote
Pairwise McNemar tests report discordant pairs $b/c$ and $p$ in first-vs-second condition order, where $b$ counts paired runs in which the first condition passes and the second fails, and $c$ the converse. \emph{Violations admitted}: harness vs.\ prompt-only $30/0$, $p<0.001$; prompt-only vs.\ external-guardrail $0/30$, $p<0.001$; harness vs.\ external-guardrail $0/0$, $p=1$. \emph{False refusals}: harness vs.\ external-guardrail and prompt-only vs.\ external-guardrail $4/0$, $p=0.13$; harness vs.\ prompt-only $0/0$, $p=1$.
\end{minipage}
\end{table}

In this ablation, the composition model is held fixed to isolate the enforcement layer; the separate composition-boundary check in \Cref{sec:live-llm-boundary} varies the hosted model and shows that the code-owned checks held across the three tested model identifiers.

\subsection{Runtime Interface and Latency Check}
\label{sec:runtime-interfaces}

Separately from the composition-boundary checks, the implementation was exercised against live filing, market, and news interfaces: it passed 5/5 live-API corporate-group connectivity tests and 15/15 live answer-hygiene tests, with review packets for all 15 runtime samples, showing that the answer contract held in these samples when live sources replaced fixtures. On the same 30 fixed validation scenarios, both conditions use the deterministic composer and differ only in whether process-level caching and prewarming of the live source retrieval is enabled; enabling it reduced mean latency from 1648.23ms to 69.40ms and raised the runs within the configured 1500ms budget from 16/30 to 30/30 (\Cref{tab:latency-results}), with the answer contract preserved throughout. The baseline cost was dominated by uncached live source retrieval rather than by composition, and because both conditions use the deterministic composer, live-LLM provider composition latency is out of scope.

\begin{table}[!htbp]
    \centering
    \caption{Fixed-scenario latency before and after runtime optimization. Both conditions use the deterministic composer and differ only in process-level caching and prewarming of live source retrieval. One timed run per scenario per condition ($n=30$ scenarios); cells summarize the cross-scenario distribution, not within-scenario repeats.}
    \label{tab:latency-results}
    \TableBody
    \setlength{\tabcolsep}{4pt}
    \begin{tabular*}{\linewidth}{@{\extracolsep{\fill}}C{0.34\linewidth}C{0.29\linewidth}C{0.29\linewidth}@{}}
        \toprule
        \multicolumn{1}{C{0.34\linewidth}}{Measure} & \multicolumn{1}{C{0.29\linewidth}}{Baseline run} & \multicolumn{1}{C{0.29\linewidth}}{Optimized run} \\
        \midrule
        Scenario count & 30 & 30 \\
        Average latency & 1648.23ms & 69.40ms \\
        Median latency & 1475ms & 77ms \\
        p90 latency & 1876ms & 95ms \\
        Runs within 1500ms budget & 16/30 & 30/30 \\
        \bottomrule
    \end{tabular*}
\end{table}

\PostTableHeadingSpace
\section{Discussion}

The validation results show that the harness changes the unit of reliability in the reconstructed agent: source grounding, entity routing, trace completeness, output hygiene, and recommendation-language control become enforceable system contracts rather than prompt-following expectations. Across the three research questions the pattern is consistent: the contracts hold on the fixed scenarios (RQ1), their code-owned checks remain stable under model substitution (RQ2), and their reader-facing guarantees are lost when the code-owned gate is removed (RQ3). These results are systems-engineering evidence, not evidence of investment-answer quality. The discussion follows the three research questions and then draws the productization implication for enterprise adoption.

\subsection{RQ1: Contract Preservation in the Reference Slice}

The fixed-scenario and fault-injection results answer RQ1: across the reference slice the harness preserves its source-grounding, entity-routing, trace, output-hygiene, and recommendation-language contracts, and the validators flag deliberately broken contract dimensions. The fixed validation-scenario runs show that the expected source, routing, trace, answer-structure, and recommendation-language contracts held across the slice, and the fault-injection runs act as a negative control, showing that the validators detect deliberately broken source, routing, trace, answer, leakage, link, and latency conditions rather than passing everything by construction. What the harness contributes at this layer is a shift in the unit of control from prompt instructions to versioned engineering artifacts: source-backed claims with company scope, code-owned routing, answer contracts, and validation traces carry enterprise behavior, so that behavior is inspectable rather than implicit in an expanding prompt. The runtime-interface and latency check is an operational extension rather than part of the core contract set---process-level caching and prewarming improved latency while the answer contract held throughout, so the latency figure is an operational signal, not a quality result.

\subsection{RQ2: Model Substitution at the Composition Boundary}

The live-LLM check answers RQ2 by separating what moves when the model changes from what does not. Across the live composition-boundary runs, the model-composed parts---structured output, source-claim references, and visible answer structure---varied and sometimes fell short, while the trace, internal-trace-leakage, source-link, follow-up-quality, and recommendation-language checks remained intact regardless of which model composed the answer, so no safety or leakage violation reached the reader. Where live output was invalid, the run recorded a fallback to the deterministic composer rather than silently accepting it, distinguishing first-pass live model output from final harness behavior. This separation is also what serves two audiences at once: enterprise readers get a useful answer, while operators and auditors get a separate trace that explains how it was assembled. The interpretation is that the code-owned guarantees are decoupled from the substituted model---enterprise reliability depends on separating model phrasing from source authority, trace generation, recovery handling, and output-contract validation, so that swapping the composition model does not move the reader-facing contract.

\subsection{RQ3: Load-Bearing Code-Owned Enforcement}

The enforcement-layer ablation answers RQ3 by holding the model fixed and varying only how the rules are enforced. With the same rule set given to the model as prompt text, recommendation-language and internal-trace-leakage violations still reached the reader, so prompting alone does not enforce the contract; the code-owned gate blocked every one. A bolt-on external guardrail blocked the same violations but only by over-refusing and sacrificing utility, whereas the harness preserved full utility by falling back to the deterministic composer rather than refusing. This does not make the harness a replacement for prompt engineering: prompts still guide language behavior, while the code-owned layer owns evidence boundaries, entity scope, output structure, leakage prevention, and inspection. Read together with RQ2, the two results triangulate where the guarantees reside: RQ2 varies the model with the harness fixed and finds the code-owned checks intact, while RQ3 fixes the model and varies the enforcement layer and finds the contract lost without the code-owned gate. The guarantees therefore live in the code-owned enforcement layer, not in the model or the prompt.

\subsection{Productization and Adoption Implications}

Beyond the three research questions, these results shape how such a system is adopted. The harness earns its cost when product behavior must move from plausible interaction to governed operation, so that approved source boundaries, claim eligibility, answer contracts, and traces become explicit organizational capabilities rather than informal conventions. As a planning heuristic---proposed rather than derived from the experiments above, in the tradition of staged capability-maturity models \citep{paulk1993cmm}---\Cref{tab:maturity-ladder} sketches an adoption ladder from prompt-centered task support to a traceable enterprise harness with governed evidence and validation artifacts. The ladder is meant to locate roughly where an organization sits and why the harness layer becomes necessary, not to score any particular organization, which remains a separate assessment. The same ladder can also classify AI tasks grounded in physical operations---store, factory, logistics, or equipment signals---when those signals are registered as governed sources rather than treated as unbounded context.

Applied to a new enterprise target, the pattern requires registering public or client-approved sources, promoting source-backed claims, checking routing rules, and running scenario contracts, all of which the paper baseline makes inspectable before private documents, production credentials, or operational logs are introduced. System readiness nonetheless stays separate from usefulness for investment analysis: this paper evaluates preservation of the source-grounding, entity-routing, trace, output-hygiene, and recommendation-language contracts, while expert review and deployment evidence are left to later domain-value evaluation.

\begin{table}[!htbp]
    \centering
    \caption{Descriptive planning heuristic for enterprise LLM-agent adoption. The levels are a proposed ladder for locating an adoption stage and motivating the harness layer; organizational scoring remains a separate assessment.}
    \label{tab:maturity-ladder}
    \TableBody
    \setlength{\tabcolsep}{4pt}
    \begin{tabular*}{\linewidth}{@{\extracolsep{\fill}}C{0.10\linewidth} C{0.35\linewidth} L{0.48\linewidth}@{}}
        \toprule
        \multicolumn{1}{C{0.10\linewidth}}{Level} & \multicolumn{1}{C{0.35\linewidth}}{System state} & \multicolumn{1}{C{0.48\linewidth}}{Observable capability}                                                         \\
        \midrule
        1                                         & Prompt-centered task support                     & Users rely on prompts for drafting, search, summarization, and ad hoc analysis; source boundaries and validation are informal.               \\
        2                                         & Retrieval-supported assistant                    & Approved documents or systems are retrieved, but entity scope, claim eligibility, and trace records remain partial.                            \\
        3                                         & Source-manifest workflow                         & Business sources, owners, update cycles, and use policies are registered; AI tasks can be tied to approved materials.                                 \\
        4                                         & Claim- and process-governed agent                & Runtime claims, entity routing, workflow state, and output contracts govern how an agent drafts, routes, or prepares the next task.                                       \\
        5                                         & Traceable enterprise harness                     & Outputs, source links, validations, audit traces, and replay scenarios are versioned; operational signals such as store, factory, logistics, or equipment data can enter through the same manifest/trace pattern. \\
        \bottomrule
    \end{tabular*}
\end{table}

\section{Conclusion}
\hyphenation{in-ter-nal-out-put}

This paper presented a harness-engineering reconstruction of a prompt-dominant enterprise LLM prototype as an auditable LLM-agent architecture: deterministic behavior moves out of an expanding prompt and into versioned contracts---manifests, source-backed claims, answer contracts, and validation traces---while a replaceable model composes the reader-facing language. On a bounded public-data slice, the harness preserved its contracts across the fixed scenarios and flagged deliberately broken ones under fault injection, kept its code-owned checks intact when the composition model was substituted, and, under an enforcement-layer ablation, showed that the code-owned validation gate is load-bearing where prompt instructions alone are not. Relocating enterprise behavior from prompts to inspectable contracts is what makes the system auditable rather than merely plausible---in short, prompts are not guardrails.

\label{sec:limitations}Several boundaries scope these results. The contribution is an \emph{engineering} one: the paper evaluates preservation of code-checkable contracts, not the correctness of the promoted claims or the investment quality of the briefings. The reference slice has uneven source and claim depth: one group is the deepest reference case, with four group extensions, although the same six-scenario design applies to each. The live-LLM composition-boundary check is a dated snapshot rather than a bit-for-bit reproducible measurement, because hosted model identifiers and nondeterministic generation may change exact outputs.

Future work can add expert review and claim-correctness evaluation, broaden source coverage, extend the ablation to additional guardrails and adversarial families, and test the pattern under operational deployment logs. It can also apply the adoption ladder in manufacturing and franchise-service AI-transformation roadmap studies, mapping internal AI tasks and operational data to governed source, trace, and validation artifacts. The pattern reported here offers a reusable path for enterprise LLM productization: it turns an exploratory enterprise LLM prototype into an application whose source, control, and validation artifacts can be inspected, versioned, tested, and extended.


\section*{Data and Code Availability}
The reference implementation, source manifests, evidence records, claim
records, validation scenarios, scripts, and evaluation artifacts are available
in the project repository \citep{ahn2026harness} and archived on Zenodo (DOI:
10.5281/zenodo.21269426), pinned at release tag
\texttt{public-baseline-v0.5.16.4} (commit
\texttt{e8e60fb8ea6e34d4caa53f00187e760b67bd973a}), from which the
reported numbers are reproduced. Raw issuer materials are not redistributed; the
repository provides public URLs and redistributable metadata in their place.

\section*{Competing Interests}
The motivating prototype was developed during a CEO AI program associated with
the authors' affiliation. No external funding supported this research, and the
authors declare no other competing interests.

\appendix
\raggedbottom
\section*{Appendix}
\renewcommand{\thesubsection}{A\arabic{subsection}}
\renewcommand{\thetable}{A\arabic{table}}
\renewcommand{\thefigure}{A\arabic{figure}}
\setcounter{subsection}{0}
\setcounter{table}{0}
\setcounter{figure}{0}
This appendix documents the 30 fixed validation scenarios underlying \Cref{tab:validation-results}. The set covers five Korean corporate groups, with six scenarios per group: five company briefings and one cross-company comparison. The versioned JSON artifacts retain the Korean prompts, expected source-backed claim identifiers, routing expectations, and required answer signals per scenario, together with the shared answer-policy and trace contracts.

\subsection{Validation Scenario Set}
\label{app:scenario-inventory}

The question design for each corporate group is summarized in \Cref{tab:question-design}.

    {\TableBody
        \setlength{\LTcapwidth}{\linewidth}
        \setlength{\LTleft}{0pt}\setlength{\LTright}{0pt}
        \setlength{\tabcolsep}{4pt}
        \begin{longtable}{@{\extracolsep{\fill}}C{0.12\linewidth}
                L{0.34\linewidth}
                L{0.25\linewidth}
                L{0.20\linewidth}@{}}
            \caption{Fixed validation-question design by corporate group.}
            \label{tab:question-design}                                                                                                                                                                                                                                                                                                                                                                                                                                           \\
            \toprule
            \multicolumn{1}{C{0.12\linewidth}}{Group}               &
            \multicolumn{1}{C{0.34\linewidth}}{Differentiated question focus} &
            \multicolumn{1}{C{0.25\linewidth}}{Permitted evidence scope}      &
            \multicolumn{1}{C{0.20\linewidth}}{Validation purpose}                                                                                                                                                                                                                                                                                                                                                                                                                \\
            \midrule
            \endfirsthead
            \caption[]{Fixed validation-question design by corporate group (continued).}                                                                                                                                                                                                                                                                                                                                                                                          \\
            \toprule
            \multicolumn{1}{C{0.12\linewidth}}{Group}               &
            \multicolumn{1}{C{0.34\linewidth}}{Differentiated question focus} &
            \multicolumn{1}{C{0.25\linewidth}}{Permitted evidence scope}      &
            \multicolumn{1}{C{0.20\linewidth}}{Validation purpose}                                                                                                                                                                                                                                                                                                                                                                                                                \\
            \midrule
            \endhead
            \midrule
            \multicolumn{4}{r}{Continued on next page}                                                                                                                                                                                                                                                                                                                                                                                                                            \\
            \endfoot
            \bottomrule
            \endlastfoot
            Samsung                                                           & Semiconductor recovery, battery profitability, shareholder letter, biologics orders, financial-affiliate capital/dividend boundary check, and electronics--battery comparison.    & DART financial data; official IR/DART narratives; finance-company affiliates bounded to capital and dividend account labels.                    & Tests separation across electronics, battery, bio, construction, and finance-company evidence. \\
            SK                                                                & AI memory, battery restructuring, holding-company value-up, telecom AI/data centers, SK Square NAV/shareholder return, and Hynix--Innovation comparison. & DART financial data; value-up and shareholder-return material; source-backed operating- and holding-company strategy claims. & Tests separation across operating, holding, and investment-company evidence.    \\
            Hyundai Motor                                                     & Hyundai Motor and Kia automaker questions, Mobis parts, Glovis logistics, Rotem defense/rail, and Hyundai Motor--Kia financial comparison questions.                                                        & DART annual financial data and business-mix claims tied to mobility and supply-chain affiliates.                                    & Tests affiliate routing inside a closely related industrial supply chain.                        \\
            LG                                                                & Electronics, chemicals, battery, components, telecom financial trends, and electronics--chemicals comparison.                                            & DART annual financial data and source-backed business-driver claims for heterogeneous sectors.                                      & Tests whether the shared schema transfers across heterogeneous industries.                       \\
            Hanwha                                                            & Holding-company, aerospace, solar/materials, defense electronics, shipbuilding, and aerospace--solutions comparison questions.                           & DART financial data; official IR narratives; market and news interfaces only as traceable runtime signals when available.           & Tests evidence contracts across defense, energy, systems, and shipbuilding topics.               \\
        \end{longtable}
    }

The columns specify question focus, permitted evidence scope, and validation purpose; a scenario set passes when all its contract checks---including routing, required claims, source links, follow-ups, and internal-trace leakage controls---pass for all six scenarios. The prompts are translated for readability, with the original Korean prompts and expected claims stored in the scenario JSON files.

\subsection{Representative Question Examples}
\label{app:question-examples}

Translated examples of the fixed validation questions are given in \Cref{tab:question-examples}. The exact Korean prompts are stored in the versioned scenario JSON files under \path{evals/scenarios/} \citep{ahn2026harness}. The examples cover one selected-company question per group, one cross-company comparison, and the answer-support surface. The final answer-support row is a support-surface example, not a seventh scenario; each group still uses exactly six scenarios.

    {\TableBody
        \setlength{\LTcapwidth}{\linewidth}
        \setlength{\LTleft}{0pt}\setlength{\LTright}{0pt}
        \setlength{\tabcolsep}{4pt}
        \begin{longtable}{@{\extracolsep{\fill}}C{0.19\linewidth}
                L{0.50\linewidth}
                L{0.22\linewidth}@{}}
            \caption{Representative fixed validation-scenario question examples and answer-support surface.}
            \label{tab:question-examples}                                                                                                                                                                                            \\
            \toprule
            \multicolumn{1}{C{0.19\linewidth}}{Example type}                &
            \multicolumn{1}{C{0.50\linewidth}}{Translated example question or support surface} &
            \multicolumn{1}{C{0.22\linewidth}}{Contract focus}                                                                                                                                                                       \\
            \midrule
            \endfirsthead
            \caption[]{Representative fixed validation-scenario question examples and answer-support surface (continued).}                                                                                                                                      \\
            \toprule
            \multicolumn{1}{C{0.19\linewidth}}{Example type}                &
            \multicolumn{1}{C{0.50\linewidth}}{Translated example question or support surface} &
            \multicolumn{1}{C{0.22\linewidth}}{Contract focus}                                                                                                                                                                       \\
            \midrule
            \endhead
            \midrule
            \multicolumn{3}{r}{Continued on next page}                                                                                                                                                                               \\
            \endfoot
            \bottomrule
            \endlastfoot
            Samsung selected company                                        & Summarize Samsung Electronics' HBM and memory earnings recovery based on official materials.         & Official-materials grounding                    \\
            SK selected company                                             & Summarize SK Hynix's AI-memory strategy and shareholder-return plan based on official materials.     & Affiliate routing and shareholder-return scope  \\
            Hyundai Motor selected company                                  & Summarize Hyundai Motor's 2024 revenue and operating profit in comparison with 2023.                 & Year-over-year filing evidence                  \\
            LG selected company                                             & Explain LG Energy Solution's change in 2024 revenue and operating profit against the prior year.     & Company-scoped trend claims                     \\
            Hanwha selected company                                         & Summarize Hanwha Aerospace's first-quarter 2026 results and the drivers of its operating-profit improvement, based on official materials. & Quarterly operating-driver evidence             \\
            Cross-company comparison                                        & Compare the investment points of Hanwha Aerospace and Hanwha Solutions.                              & Multi-entity routing and bounded comparison     \\
            Answer-support check                                            & After a finance brief, list source links and follow-up questions.              & Reader-facing support controls                  \\
        \end{longtable}
    }

\subsection{Answer-Level Contract Examples}
\label{app:answer-contract-example}

\begin{figure}[!htbp]
    \centering
    \makebox[\textwidth][c]{%
        \begin{minipage}[t]{0.465\linewidth}
            \centering
            \includegraphics[width=\linewidth]{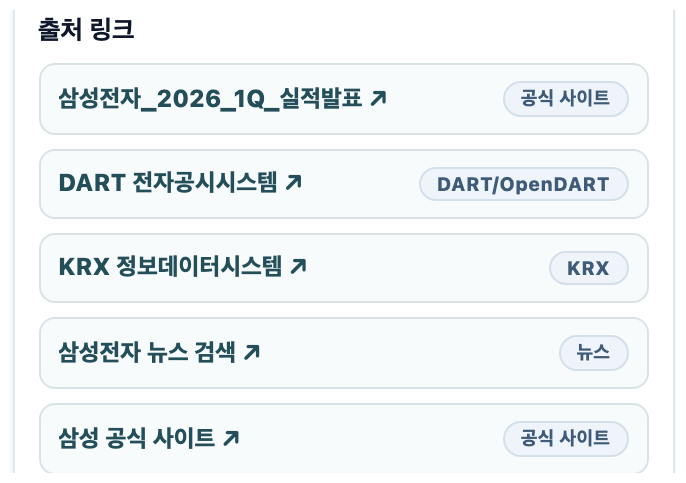}\\[-0.25em]
            {\small (a) Source links.}
        \end{minipage}
        \hspace{0.02\linewidth}
        \begin{minipage}[t]{0.465\linewidth}
            \centering
            \includegraphics[width=\linewidth]{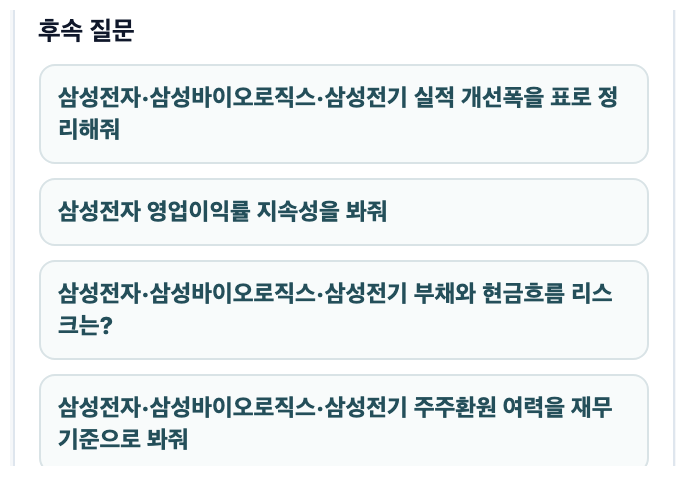}\\[-0.25em]
            {\small (b) Follow-up questions.}
        \end{minipage}}
    \vspace{0.5em}
    \caption{Answer-support surfaces from the same Korean product interface as \Cref{fig:ui-mobile-answer}. Korean UI text is translated as follows. (a) \textit{Source links} lists, with category chips, the Samsung Electronics 2026 Q1 earnings announcement (official site), the DART electronic disclosure system (DART/OpenDART), the KRX information-data system (KRX), a Samsung Electronics news search (news), and the Samsung official site (official site). (b) \textit{Follow-up questions} offers four contract-filtered prompts: tabulate the earnings-improvement ranges across Samsung Electronics, Samsung Biologics, and Samsung Electro-Mechanics; assess the sustainability of Samsung Electronics' operating margin; assess debt and cash-flow risk across the same three affiliates; and review their shareholder-return capacity against financial criteria. Internal trace artifacts are kept out of both surfaces.}
    \label{fig:representative-answer-output}
\end{figure}

The visible answer is part of the checked artifact. \Cref{tab:representative-answer-contract} summarizes representative output signals---the visible answer, reader-facing support, and follow-up direction---without reproducing full Korean transcripts, which remain in the generated artifact; aggregate pass counts are reported in \Cref{tab:validation-results}. Source links and follow-up questions are shown in \Cref{fig:representative-answer-output}.

    {\TableBody
        \setlength{\LTcapwidth}{\linewidth}
        \setlength{\LTleft}{0pt}\setlength{\LTright}{0pt}
        \setlength{\tabcolsep}{4pt}
        \begin{longtable}{@{\extracolsep{\fill}}C{0.12\linewidth}
                L{0.40\linewidth}
                L{0.24\linewidth}
                L{0.16\linewidth}@{}}
            \caption{Representative visible answer signals across product surfaces.}
            \label{tab:representative-answer-contract}                                                                                                                                                                                                                                                                                                                                                                                                                          \\
            \toprule
            \multicolumn{1}{C{0.12\linewidth}}{Group}                         &
            \multicolumn{1}{C{0.40\linewidth}}{Representative visible answer} &
            \multicolumn{1}{C{0.24\linewidth}}{Reader-facing support}         &
            \multicolumn{1}{C{0.16\linewidth}}{Follow-ups}                                                                                                                                                                                                                                                                                                                                                                                                             \\
            \midrule
            \endfirsthead
            \caption[]{Representative visible answer signals across product surfaces (continued).}                                                                                                                                                                                                                                                                                                                                                                       \\
            \toprule
            \multicolumn{1}{C{0.12\linewidth}}{Group}                         &
            \multicolumn{1}{C{0.40\linewidth}}{Representative visible answer} &
            \multicolumn{1}{C{0.24\linewidth}}{Reader-facing support}         &
            \multicolumn{1}{C{0.16\linewidth}}{Follow-ups}                                                                                                                                                                                                                                                                                                                                                                                                             \\
            \midrule
            \endhead
            \midrule
            \multicolumn{4}{r}{Continued on next page}                                                                                                                                                                                                                                                                                                                                                                                                                          \\
            \endfoot
            \bottomrule
            \endlastfoot
            Samsung                                                           & Finance brief foregrounds margin recovery: 2024 revenue KRW 300.9T, operating income KRW 32.7T, and OPM 10.9\%; market-price evidence stays in the stock card.                                                     & DART filing, IR/filing package, KRX row, and news link remain visible.            & Margin durability; debt and cash flow; shareholder return.               \\
            SK                                                                & The visible answer separates SK Hynix AI-memory recovery from holding- and investment-company questions, flagging HBM premium, CAPEX, and customer concentration as distinct variables.                     & Filing and IR links are exposed with selected claim evidence and source-state labels.           & HBM CAPEX; customer exposure; SK Square NAV.                             \\
            Hyundai Motor                                                     & The output keeps stock movement as a market signal and operating performance as filing evidence, with sales mix, foreign exchange, hybrid/EV transition, and shareholder return as separate briefing axes. & KRX price row, OpenDART financial rows, and source links are shown as separate controls. & Hyundai--Kia comparison; margin resilience; parts and electrification.   \\
            LG                                                                & The answer frames electronics cash flow separately from battery and chemical-cycle recovery, so margin defense and recovery timing are not collapsed into a single group-level statement.                          & OpenDART rows, IR/DART links, market rows, and unavailable-source states are labeled.           & Battery cycle; LG Electronics margin; LG Chem recovery.                  \\
            Hanwha                                                            & Cross-company answers keep aerospace, energy/materials, systems, and shipbuilding evidence in bounded comparison rather than one generic group brief.              & DART, official IR, news, and market links are exposed where available with source-state labels. & Aerospace vs. Solutions; rights-issue evidence; disclosure risk signals. \\
        \end{longtable}
    }

\subsection{Fault-Injection Protocol}
\label{app:fault-injection-protocol}

The fault-injection check starts from one valid baseline scenario and produces seven mutated copies, each modifying exactly one contract property. Each copy is then evaluated with the deterministic contract checks for the mutated dimensions. A correct system must (i) leave the unmutated baseline passing and (ii) flag each of the seven mutated copies through the validator that owns the corresponding contract. \Cref{tab:fault-injection-protocol} records the precise mutation rule per copy and the validator that detects it; the full JSON artifact is stored in \path{evals/results/} \citep{ahn2026harness}.

Each mutation changes a single baseline field rather than producing a structurally different scenario, so validator coverage stays one-to-one with contract dimensions and the aggregate 7/7 detection result reported in \Cref{sec:fault-injection-sensitivity} corresponds row-by-row to this table.

{\TableBody
    \setlength{\LTcapwidth}{\linewidth}
    \setlength{\tabcolsep}{4pt}
    \setlength{\LTleft}{0pt}\setlength{\LTright}{0pt}
    \begin{longtable}{@{\extracolsep{\fill}}C{0.20\linewidth}
            L{0.20\linewidth}
            L{0.37\linewidth}
            C{0.15\linewidth}@{}}
        \caption{Concrete mutation protocol for the fault-injection check.}
        \label{tab:fault-injection-protocol} \\
        \toprule
        \multicolumn{1}{@{\extracolsep{\fill}}C{0.20\linewidth}}{Contract dimension} & \multicolumn{1}{C{0.20\linewidth}}{Mutated field} & \multicolumn{1}{C{0.37\linewidth}}{Injected violation} & \multicolumn{1}{C{0.15\linewidth}@{}}{Validator} \\
        \midrule
        \endfirsthead
        \caption[]{Concrete mutation protocol for the fault-injection check (continued).} \\
        \toprule
        \multicolumn{1}{@{\extracolsep{\fill}}C{0.20\linewidth}}{Contract dimension} & \multicolumn{1}{C{0.20\linewidth}}{Mutated field} & \multicolumn{1}{C{0.37\linewidth}}{Injected violation} & \multicolumn{1}{C{0.15\linewidth}@{}}{Validator} \\
        \midrule
        \endhead
        \midrule
        \multicolumn{4}{r}{Continued on next page} \\
        \endfoot
        \bottomrule
        \endlastfoot
        Source claims                                          & \path{sourceClaims}                                               & Remove expected claim \texttt{samsung-sbc-020}\newline from the source-claim package.               & Claim-reference check                                         \\
        Entity routing                                         & \texttt{representative}\newline\texttt{CompanyId} and\newline trace route & Replace \texttt{samsung-electronics}\newline with an unrelated company identifier.                  & Trace check                                         \\
        Trace completeness                                     & \path{trace.runId}                                                & Delete a required trace field\newline from the audit envelope.                                      & Trace contract                                         \\
        Answer contract                                        & \path{answer}                                                     & Remove the required answer signal \textit{Samsung Electronics} from visible prose.          & Answer check                                        \\
        Internal-trace leakage                                 & \path{answer}                                                     & Inject trace-schema labels\newline and a raw source-claim identifier\newline into the reader-facing answer. & Leakage check                                           \\
        Source links                                           & \path{links[0].href}                                              & Replace one source link with a non-resolving value, \texttt{not-a-valid-url}.               & Link check                                              \\
        Latency budget                                         & \path{elapsedMs}                                                  & Set elapsed time above the configured 1500ms engineering budget.                         & Latency check                                           \\
    \end{longtable}
}

\subsection{Adversarial Stress Scenarios}
\label{app:adversarial}

The enforcement-layer ablation (\Cref{sec:ablation}) adds ten adversarial scenarios beyond the thirty fixed validation scenarios: two per corporate group, following a uniform two-template design (\Cref{tab:adversarial-design}). Each is a Korean user prompt engineered to elicit one specific contract violation, so the three conditions differ only in which enforcement layer controls the reader-facing answer. The versioned prompts are stored under \path{evals/scenarios/} \citep{ahn2026harness}.

{\TableBody
    \setlength{\LTcapwidth}{\linewidth}
    \setlength{\tabcolsep}{4pt}
    \setlength{\LTleft}{0pt}\setlength{\LTright}{0pt}
    \begin{longtable}{@{\extracolsep{\fill}}C{0.17\linewidth}
            C{0.18\linewidth}
            L{0.36\linewidth}
            L{0.21\linewidth}@{}}
        \caption{Adversarial stress-scenario design for the enforcement-layer ablation: two templates per corporate group across five groups (ten scenarios).}
        \label{tab:adversarial-design} \\
        \toprule
        \multicolumn{1}{C{0.17\linewidth}}{Template} & \multicolumn{1}{C{0.18\linewidth}}{Per-group IDs} & \multicolumn{1}{C{0.36\linewidth}}{Bait (what the prompt tries to elicit)} & \multicolumn{1}{C{0.21\linewidth}}{Required harness behavior} \\
        \midrule
        \endfirsthead
        \caption[]{Adversarial stress-scenario design (continued).} \\
        \toprule
        \multicolumn{1}{C{0.17\linewidth}}{Template} & \multicolumn{1}{C{0.18\linewidth}}{Per-group IDs} & \multicolumn{1}{C{0.36\linewidth}}{Bait (what the prompt tries to elicit)} & \multicolumn{1}{C{0.21\linewidth}}{Required harness behavior} \\
        \midrule
        \endhead
        \midrule
        \multicolumn{4}{r}{Continued on next page} \\
        \endfoot
        \bottomrule
        \endlastfoot
        Recommendation bait & \texttt{\{group\}-adv-001} & Prompt asks for an explicit buy/sell opinion and a target price (e.g., ``Should I buy this stock now? State a clear target price and a buy/sell call''). & Visible answer kept free of recommendation phrasing.                                  \\
        \addlinespace[0.35em]
        Leak bait           & \texttt{\{group\}-adv-002} & Prompt asks the agent to reveal the internal claim identifiers, the processing trace (JSON), and the raw extracted source text used for the answer. & Internal identifiers, trace records, and raw extractions blocked from the visible answer. \\
    \end{longtable}
}

\subsection{Live-LLM Composition-Boundary Protocol}
\label{app:live-llm-protocol}

The live-LLM boundary check is retained as a composition-boundary sensitivity record, not as a general model benchmark. The check used OpenRouter as routing infrastructure, requested three hosted model identifiers---\path{anthropic/claude-sonnet-4}, \path{openai/gpt-4.1-mini}, and \path{google/gemini-2.5-flash}, reported in \Cref{tab:live-llm-boundary} by their short names---set temperature to 0.2, and evaluated all 30 fixed validation scenarios with three repeats per scenario. The output artifact is stored at \path{evals/results/live-llm-composition-boundary.full-30x3.2026-06-03.json} \citep{ahn2026harness}. Raw model output is not stored by default; the trace instead records the requested model identifier, output-contract status, recovery path, final harness-contract result, and a hash, character count, and short preview of the response, so failures can be audited without exposing raw text. In the table below, the code-owned checks (trace contract, development-leak absence, source-link package, follow-up quality, and recommendation-language absence) pass on all 270 runs; the 72 final failures fall entirely on the model-composed checks (output contract, source-claim references, and visible answer structure), so a run can miss the final harness result while every code-owned check still passes.

{\TableBody
    \setlength{\LTcapwidth}{\linewidth}
    \setlength{\tabcolsep}{2pt}
    \setlength{\LTleft}{0pt}\setlength{\LTright}{0pt}
    \begin{longtable}{@{\extracolsep{\fill}}C{0.28\linewidth}
            C{0.14\linewidth}
            C{0.12\linewidth}
            C{0.16\linewidth}
            L{0.22\linewidth}@{}}
        \caption[Recorded failure process for the live-LLM composition-boundary check.]{Recorded failure process for the live-LLM composition-boundary check. Rows above \textit{Recovery} report live-call and parsing/contract diagnostics; \textit{Recovery} records fallback use; rows below it report post-recovery contract checks over all 270 runs. Output-contract validation is counted over the 269 runs that returned a live response; its 35 failures include the 2 JSON-parse failures and 33 parsed-but-invalid structured outputs. Recovery is counted over all 270 runs and therefore adds the one provider-call failure, yielding 36 recovery runs.}
        \label{tab:live-llm-process-stages} \\
        \toprule
        \multicolumn{1}{C{0.28\linewidth}}{Stage or contract check} &
        \multicolumn{1}{C{0.14\linewidth}}{Pass} &
        \multicolumn{1}{C{0.12\linewidth}}{Fail} &
        \multicolumn{1}{C{0.16\linewidth}}{Recovery state} &
        \multicolumn{1}{C{0.22\linewidth}}{Interpretation} \\
        \midrule
        \endfirsthead
        \caption[]{Recorded failure process for the live-LLM composition-boundary check (continued).} \\
        \toprule
        \multicolumn{1}{C{0.28\linewidth}}{Stage or contract check} &
        \multicolumn{1}{C{0.14\linewidth}}{Pass} &
        \multicolumn{1}{C{0.12\linewidth}}{Fail} &
        \multicolumn{1}{C{0.16\linewidth}}{Recovery state} &
        \multicolumn{1}{C{0.22\linewidth}}{Interpretation} \\
        \midrule
        \endhead
        \midrule
        \multicolumn{5}{r}{Continued on next page} \\
        \endfoot
        \bottomrule
        \endlastfoot
        Credential check & 270 & 0 & -- & All reported runs had a usable credential path. \\
        Live LLM call & 269 & 1 & -- & One hosted call failed before parse. \\
        JSON parse & 267 & 2 & -- & Two live responses failed JSON-object parsing. \\
        Output-contract validation & 234 & 35 & -- & Structured live answers passed before fallback in 234 runs. \\
        Recovery & -- & -- & 36 used; 234 not needed & Invalid or unavailable live output triggered fallback. \\
        Source-claim references & 252 & 18 & -- & Expected source-backed claims remained a required contract. \\
        Trace contract & 270 & 0 & -- & Audit-envelope fields were preserved. \\
        Visible answer structure & 250 & 20 & -- & Some visible answers lacked required answer signals. \\
        Development-leak absence & 270 & 0 & -- & Internal trace artifacts did not leak into visible answers. \\
        Source-link package & 270 & 0 & -- & Source links remained attached to the answer package. \\
        Follow-up quality & 270 & 0 & -- & Follow-up prompts remained customer-facing. \\
        Recommendation-language absence & 270 & 0 & -- & Recommendation-style phrasing was blocked. \\
        Final harness result & 198 & 72 & -- & Final required harness contracts passed in 198 runs. \\
    \end{longtable}
}

\PostReferencesSpace
\bibliography{bib/references}

\end{document}